\documentclass[final,3p,times,twocolumn]{elsarticle}

\usepackage{amssymb}
\usepackage{amsmath}
\usepackage{algorithm2e}
\usepackage{subcaption}
\usepackage[colorlinks,citecolor=citecolor]{hyperref}

\usepackage{xspace}

\makeatletter
\DeclareRobustCommand\onedot{\futurelet\@let@token\@onedot}
\def\@onedot{\ifx\@let@token.\else.\null\fi\xspace}

\def\ie{\emph{i.e}\onedot}

\makeatother

\usepackage{url}

\usepackage{wrapfig}

\newcommand{\cutBeforeFigureCaption}{\vspace{0pt}}
\newcommand{\cutAfterFigureCaption}{\vspace{0pt}}
\newcommand{\cutBeforeTableCaption}{\vspace{0pt}}
\newcommand{\cutAfterTableCaption}{\vspace{0pt}}
\newcommand{\cutBeforeSubTableCaption}{\vspace{0pt}}
\newcommand{\cutAfterSubTableCaption}{\vspace{0pt}}
\newcommand{\cutBeforeSection}{\vspace{0pt}}
\newcommand{\cutAfterSection}{\vspace{0pt}}

\usepackage{xcolor}
\newcommand{\re}[1]{\textcolor{black}{#1}}

\journal{Pattern Recognition}

\begin{document}

\begin{frontmatter}



\title{Few-shot Forgery Detection via Guided Adversarial Interpolation}


\author[a1]{Haonan Qiu}
\ead{HAONAN002@e.ntu.edu.sg}

\author[a2]{Siyu Chen}
\ead{siyuchen@pku.edu.cn}

\author[a2]{Bei Gan}
\ead{ganbei@sensetime.com}

\author[a2]{Kun Wang}
\ead{wangkun@sensetime.com}

\author[a2]{Huafeng Shi}
\ead{shihuafeng1@sensetime.com}

\author[a2]{Jing Shao}
\ead{shaojing@sensetime.com}

\author[a1]{Ziwei Liu\corref{cor1}}
\ead{ziwei.liu@ntu.edu.sg}
\cortext[cor1]{Corresponding Author: Ziwei Liu (ziwei.liu@ntu.edu.sg)}

\affiliation[a1]{organization={S-Lab, Nanyang Technological University}, addressline={50 Nanyang Ave}, postcode={639798}, country={Singapore}}

\affiliation[a2]{organization={SenseTime Research}, addressline={Hongmei Road}, city={Shanghai}, postcode={200233}, country={China}}



\begin{abstract}
   The increase in face manipulation models has led to a critical issue in society - the synthesis of realistic visual media. With the emergence of new forgery approaches at an unprecedented rate, existing forgery detection methods suffer from significant performance drops when applied to unseen novel forgery approaches. In this work, we address the few-shot forgery detection problem by \textbf{1)} designing a comprehensive benchmark based on coverage analysis among various forgery approaches, and \textbf{2)} proposing \textbf{Guided Adversarial Interpolation (GAI)}. Our key insight is that there exist transferable distribution characteristics between majority and minority forgery classes\footnote{Majority class: class with abundant samples; minority class: class with scarce samples.}. Specifically, we enhance the discriminative ability against novel forgery approaches via adversarially interpolating the forgery artifacts of the minority samples to the majority samples under the guidance of a teacher network. Unlike the standard re-balancing method which usually results in over-fitting to minority classes, our method simultaneously takes account of the diversity of majority information as well as the significance of minority information. Extensive experiments demonstrate that our GAI achieves state-of-the-art performances on the established few-shot forgery detection benchmark.
   Notably, our method is also validated to be robust to choices of majority and minority forgery approaches. 
   The formal publication version is available in \href{https://doi.org/10.1016/j.patcog.2023.109863}{Pattern Recognition}.
\end{abstract}



\begin{keyword}
Forgery Detection \sep DeepFake \sep Few-shot \sep Face Manipulation


\end{keyword}

\end{frontmatter}


\cutBeforeSection
\section{Introduction}
\cutAfterSection
\label{sec:intro}

\begin{figure*}[!ht]
\begin{center}
\includegraphics[width=0.9\linewidth]{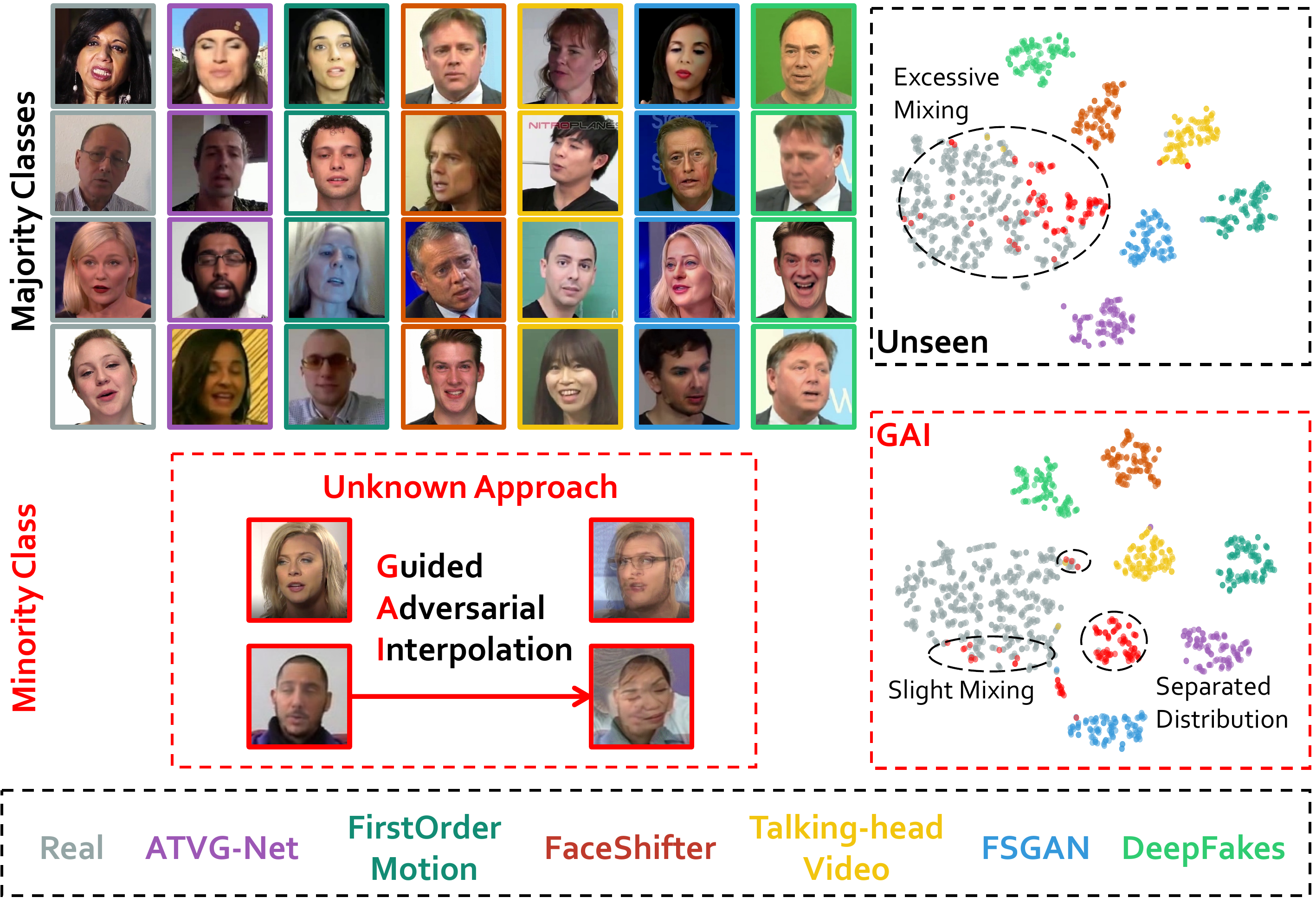}
\end{center}
\cutBeforeFigureCaption
\caption{\textbf{Problem Motivation.}
Current forgery detection methods are prone to failing at an unknown forgery approach.
Therefore, at least a few samples of the new forgery approach are needed, and should be fully utilized to achieve high accuracy.
}
\cutAfterFigureCaption
\label{fig:teaser}
\end{figure*}

With the development of computer vision and graphics, synthesized images, especially forged face images, have been extremely difficult for humans to distinguish. 
Popular face-swap approaches can allow a person's face to be naturally swapped into another irrelevant video, which poses a significant risk of exploitation.
Typically, current forgery detection methods involve training a specific detector on a predetermined dataset~\cite{jiang2020deeperforensics}, where forged data are generated using previously available forgery approaches.
Such a detector is only designed to detect specific forgery approaches that were present in the training set. Therefore, it is likely that the forgery detection method will fail to identify media that has been generated by a new, previously unseen approach~\cite{yang2020one}.
Based on Fig.~\ref{fig:teaser}, we can observe that samples obtained from the unseen forgery approach have a distribution that is similar to real data when assessed with a naively trained detection model.

An ideal forgery detection system would be capable of identifying any forged samples, even those previously unseen, without relying on predetermined knowledge~\cite{sun2021domain}. But current technology has yet to achieve this lofty goal due to various forgery types and new emerging forgery approaches. Certain methods may work effectively on one forgery type, but fail to detect others~\cite{cozzolino2018forensictransfer}. Therefore, defenders need to update their detection models after some new forged samples succeed to attack their systems.

A conventional solution to address the challenges of detecting the new forgery approach is the addition of novel samples for model fine-tuning.
However, the amount of those novel samples is highly limited due to \textbf{1)} high reproduction cost, or even \textbf{2)} unknown implementation details.
In a common scenario of the real world, the only available novel samples for defenders are those who succeed to attack their detection system and then are collected afterward.
Therefore, few-shot forgery detection is becoming meaningful and in urgent need of being explored.
Although some works~\cite{yang2020one,sun2021domain} start to pay attention to this field, their settings miss consideration of the correlation among different forgery approaches.
For example, the dataset used in FT~\cite{cozzolino2018forensictransfer} only considers four forgery approaches that relate to face swap and face reenactment.
The limited forgery types make it hard to measure whether a detection method is robust for the new forgery approach.
Even worse, there is no public benchmark built specifically for few-shot forgery detection, making it hard to measure the performance of forgery detection methods under the few-shot setting.
To address the aforementioned issues, we put forward a new benchmark composed of four datasets based on coverage analysis among various forgery approaches. 
Specifically, we involve a total of thirteen advanced forgery approaches, including face transfer, face swap, face reenactment, and face editing. Those approaches cover almost all types of face forgery, enabling us to assess the effectiveness of forgery detection methods thoroughly. Following coverage analysis, we create several combinations of majority and minority classes for each dataset.
More details will be introduced in Sec.~\ref{sec:benchmark}.

Then we conduct a comprehensive evaluation of several advanced forgery detection methods on our benchmark, and the experimental results are unsatisfactory, highlighting the challenges of the few-shot setting. One possible intuitive solution is to extend the distribution of the minority class. But previous research suggests that over-sampling or simple data augmentation within a minority domain has not yielded satisfactory results~\cite{kim2020m2m}.

To overcome this challenge, it is essential to leverage all available information beyond the minority domain. While minority samples contain information about new forgery approaches, they lack diversity in identities due to limited numbers. Conversely, major samples contain information about various identities but do not offer any information about the new forgery approach.
In order to augment the minority domain through the majority domains, 
we propose \textbf{Guided Adversarial Interpolation (GAI)}, which enhances the distribution information of the minority domain via adversarially transferring majority samples to the minority domain under the guidance of a teacher network.
In contrast to directly adding perturbations to majority samples, we interpolate majority samples and minority samples, which enables the forgery artifacts of the minority domain to remain in the augmented samples. A tensor of interpolating coefficients is updated with the teacher's guidance, making the newly generated samples located in the minority domain.
Integrating the benefits of both types of samples, GAI effectively improves predictions on the minority domain samples following data augmentation in the minority domain while maintaining accuracy in other majority domains.

Experimental results show that our method achieves state-of-the-art performances on the few-shot forgery detection benchmark, and its performance is robust to the choices of majority and minority forgery approaches. 
The t-SNE map in Fig.~\ref{fig:teaser} also indicates our GAI is able to separate minority samples into an independent cluster, thus obtaining high prediction accuracy in terms of forgery detection.

To sum up, our contributions in this work include:
\begin{itemize}
  \item Based on forgery coverage analysis, we build four datasets and form a few-shot forgery detection benchmark. To the best of our knowledge, this is the first benchmark specifically set up for few-shot forgery detection. 
  \item We propose a concise yet effective framework, Guided Adversarial Interpolation (GAI). It interpolates the forgery artifacts of minority samples with majority samples guided by a teacher network, which is capable of integrating the benefits of both minority and majority domains.
  \item Extensive experiments demonstrate that GAI achieves state-of-the-art performance on the established benchmark. GAI is also robust to choices of majority and minority forgery approaches. 
\end{itemize}

\cutBeforeSection
\section{Related Works}
\cutAfterSection

\noindent\textbf{Forgery Synthesis and Detection.}

The easiest and most common forgery approaches are copy-move. Mostly, they are addressed by detection methods based on feature extraction and matching~\cite{BI2018161,ZHONG2022108286}. Recently, the learning-based method QDL~\cite{ARIA2022213} is proposed to detect copy-move through manipulation detection and similarity detection.

In addition, AI-based image and video synthesis algorithms ~\cite{choi2020stargan, Karras2019stylegan2, CelebAMask-HQ} have flourished thanks to recent advances in deep generative models~\cite{goodfellow2014generative}.
However, realistic forgery technologies bring hidden dangers of abuse.
To address this pressing issue, numerous forgery detection datasets~\cite{rossler2019faceforensics++,li2020celeb} have been set up, from early ones including SwapMe and FaceSwap~\cite{zhou2017two}, to recently emerged large-scale benchmarks such as DeeperForensics-1.0~\cite{jiang2020deeperforensics}, DFFD~\cite{dang2020detection} and ForgeryNet~\cite{he2021forgerynet}. Following these datasets, a large number of works have been dedicated to forgery detection, and have made remarkable progress. Some methods~\cite{mccloskey2018detecting} utilize color-space features. More recent studies~\cite{PU2022108832,SHANG2021107950} leverage deep neural networks to extract high-level representations for more accurate classification. 
\re{Considering the model generalization to unknown manipulations and scenes, SBI~\cite{shiohara2022detecting} produces general and hardly recognizable fake samples through blending pseudo source and target images from single pristine images, encouraging classifiers to learn generic and robust representations without overfitting.}
And for DeepFake videos, Chen~\cite{CHEN2023109179} proposes to detect them via watching bi-granularity artifacts.
%


\noindent\textbf{Few-Shot Classification.}
Previous works rarely consider the case of unknown forgery approaches appearing in testing, which is however the most common real-world scenario. 
Although some work~\cite{cozzolino2018forensictransfer,yang2020one,sun2021domain} research similar problems in this direction, their settings are all limited and none of them provide a public benchmark for evaluation.
LTW~\cite{sun2021domain} only considers the zero-shot case, and its accuracy is mostly around $60\%$ which is not far from random guesses and not meaningful to realistic applications.
FT~\cite{cozzolino2018forensictransfer} and OSFG~\cite{yang2020one} use samples from unknown forgery approaches but do not consider coverage among forgery approaches.
These two few-shot forgery detection methods will be evaluated and compared on our benchmark.

Regarding few-shot forgery detection as a long-tailed problem, which is very common in an open world and attracts lots of research attention~\cite{liu2019large, kim2020m2m},
FT~\cite{cozzolino2018forensictransfer} and OSFG~\cite{yang2020one} leverage domain adaptation to supply the information of few-shot classes.
Similar to our work, M2m~\cite{kim2020m2m} transfers the knowledge from the majority domain to the minority domain to expand the information of the few-shot class.
However, transferred samples from M2m are too close to the majority samples and do not help much for this few-shot forgery detection task, while its teacher-student pipeline inspires our work to use guidance during domain interpolation. 


\cutBeforeSection
\section{Few-Shot Forgery Detection Benchmark}
\cutAfterSection
\label{sec:benchmark}

In this section, we formally define the task of few-shot forgery detection, and then systematically set up a large-scale benchmark for evaluating performances on this task.

\subsection{Problem Definition}
\label{subsec:prob_def}

A basic forgery detection training dataset generally consists of both real data and forged data from $n$ forgery approaches. 
This base dataset can be represented by $\mathcal{D}^\text{train}_\text{base} = \left\{d^\text{train}_{0},d^\text{train}_{1},...,d^\text{train}_{n} \right\}$ where $d^\text{train}_{0}$ represents real data, and the other $n$ sets are forged ones. 
Detection models trained on $\mathcal{D}^\text{train}_\text{base}$ will face a challenging test dataset formed by $m+1$ domains $\mathcal{D}^\text{test} = \left\{d^\text{test}_{0},d^\text{test}_{1},...,d^\text{test}_{m} \right\}$ ($m\geq n$). The last $m-n$ domains are unseen during training.
In the real world, $m\gg n$. Yet in this paper, we focus on taking the first step and assume $m = n+1$, which implies only one novel forgery approach not appearing in $\mathcal{D}^\text{train}_\text{base}$ exists in the test set. 
As discussed in Sec.~\ref{sec:intro}, defenders may obtain a limited number of samples of the novel forgery approach for training when they find their detection system is attacked.
Therefore, the complete training dataset for few-shot forgery detection is represented by $\mathcal{D}^\text{train} = \mathcal{D}^\text{train}_\text{base} + \{d^\text{train}_{n+1}\} = \left\{d^\text{train}_{0},d^\text{train}_{1},..., d^\text{train}_{n+1} \right\}$, where $d^\text{train}_{n+1}$ contains samples from the new forgery approach.
For the subset of data with index $c$ ($c=0,1,\cdots,n+1$), $d^\text{train}_c = \left\{(\mathbf{x}_i,\mathbf{y}_i)\right\}_{i=1}^{N_c}$ consists of image-label pairs, where $\mathbf{x}_i \in \mathbb{R}^{H \times W \times D}$ and $\mathbf{y}_i=c$ denote the image and the ground-truth label, respectively.
Note that $N_{n+1}\ll N_{c^\text{major}}$ ($c^\text{major}=0,1,\cdots,n$) should be held due to the scarcity of novel class data, which is why we also name $d^\text{train}_{n+1}$ as the minority class.

The goal of few-shot forgery detection is to train on $\mathcal{D}^\text{train}$ and achieve good performance on $\mathcal{D}^\text{test}$, especially the novel domain $d^\text{test}_{n+1}$ which corresponds to the minority class $d^\text{train}_{n+1}$ in the training set.

\subsection{Few-shot Forgery Detection Datasets}

Currently, there is no public dataset built specifically for few-shot forgery detection.
To evaluate performances of previous works and our proposed GAI on this task, we build up several sets of data by combining samples generated from various forgery approaches. In order to control the difficulty of the task, we first perform cross-forgery analysis among approaches, and then select appropriate ones for the few-shot scenario according to the results.

\noindent\textbf{Forgery Data Generation.}
\label{subsubsection:data_gen}
We first construct a gigantic data base with abundant real data and a range of forgery approaches. Our few-shot forgery detection datasets are all subsets sampled from this data base. We pick four publicly available datasets~\cite{cao2014crema,livingstone2018ryerson,Chung18b,ephrat2018looking} as sources of real data, which are highly diversified.

We generate forgery data manipulated by thirteen forgery approaches including face transfer, face swap, face reenactment, and face editing.
\textbf{1)} Face transfer replaces identity-aware and
identity-agnostic content: BlendFace blends the face into the target area with Poisson fusion after the affine transformation. MMReplacement transfer the face through the 3DMM model. \textbf{2)} Face swap replaces identity only: FSGAN~\cite{nirkin2019fsgan}, DeepFakes~\cite{petrov2020deepfacelab}, FaceShifter~\cite{li2019faceshifter} swap faces based on deep neural networks. \textbf{3)} Face reenactment drives the face to do the new facial action: FirstOrderMotion~\cite{Siarohin_2019_NeurIPS},  ATVG-Net~\cite{chen2019hierarchical}, and Talking-head Video~\cite{fried2019text} drive the face by giving new video, audio, and text respectively. \textbf{4)} Face editing manipulates face attributes:  StarGAN2~\cite{choi2020stargan} edits face attributes by giving a new reference face. MaskGAN~\cite{CelebAMask-HQ} and SC-FEGAN~\cite{Jo_2019_ICCV} manipulate faces by drawing new masks and strikes respectively. DiscoFaceGAN~\cite{deng2020disentangled} modifies faces via changing disentangled latent.

Since real and manipulated data include both images and videos, we sample frames from videos to form an image-only dataset with $\sim2.4$M data. In particular, there are $\sim1.2$M real images, and the number of forgery data from each forgery approach varies from $20$k to $110$k. We split the dataset into training and test sets with the ratio $\sim14:1$, and identities in these two sets do not overlap. Compared to previous forgery detection datasets~\cite{jiang2020deeperforensics,rossler2019faceforensics++,li2020celeb,dang2020detection}, our data base excels in terms of diversity, and we provide clear annotations about forgery approaches used for data generation.

\begin{figure}[!t]
\begin{center}
\includegraphics[width=0.9\linewidth]{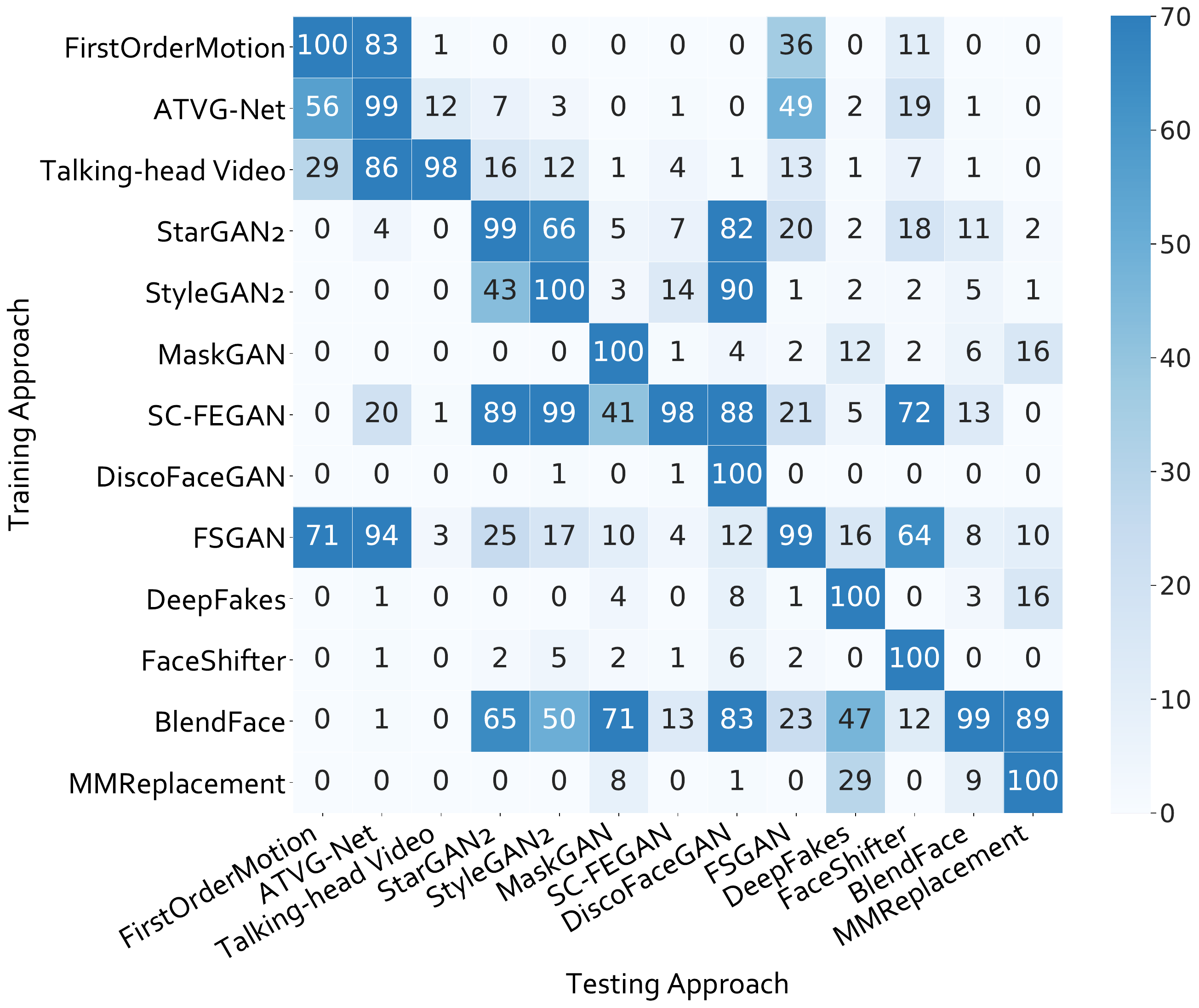}
\end{center}
\cutBeforeFigureCaption
\caption{\textbf{Coverage Analysis.} Heat map of cross-forgery accuracy with thirteen forgery approaches. Widely existing low coverage accuracy shows that a well-trained detection model often fails at unseen forgery approaches.}
\cutAfterFigureCaption
\label{fig:heat-cross}
\end{figure}

\noindent\textbf{Coverage Analysis on Forgery Approaches.}
\label{subsubsection:Coverage}
Different from other few-shot classification tasks, there exist strong correlations among some forgery approaches. Training on forgery approach A may naturally result in high accuracy on another forgery approach B~\cite{yang2020one}. To avoid easily saturating performances, we should circumvent such ``coverage" phenomena when selecting the minority class in the dataset construction process.

After forgery data generation, we test coverage abilities of our thirteen forgery approaches by training a detector (Xception~\cite{chollet2017xception}) on data of each forgery approach combined with real ones, respectively. 
As shown in Fig.~\ref{fig:heat-cross}, cross-forgery coverage performances differ drastically. 
Quite a large number of pairs of forgery approaches give nearly zero transfer accuracy, while some other pairs exhibit excellent coverage behavior ($70\%+$ accuracy).

\begin{figure}[!t]
\begin{center}
\includegraphics[width=0.9\linewidth]{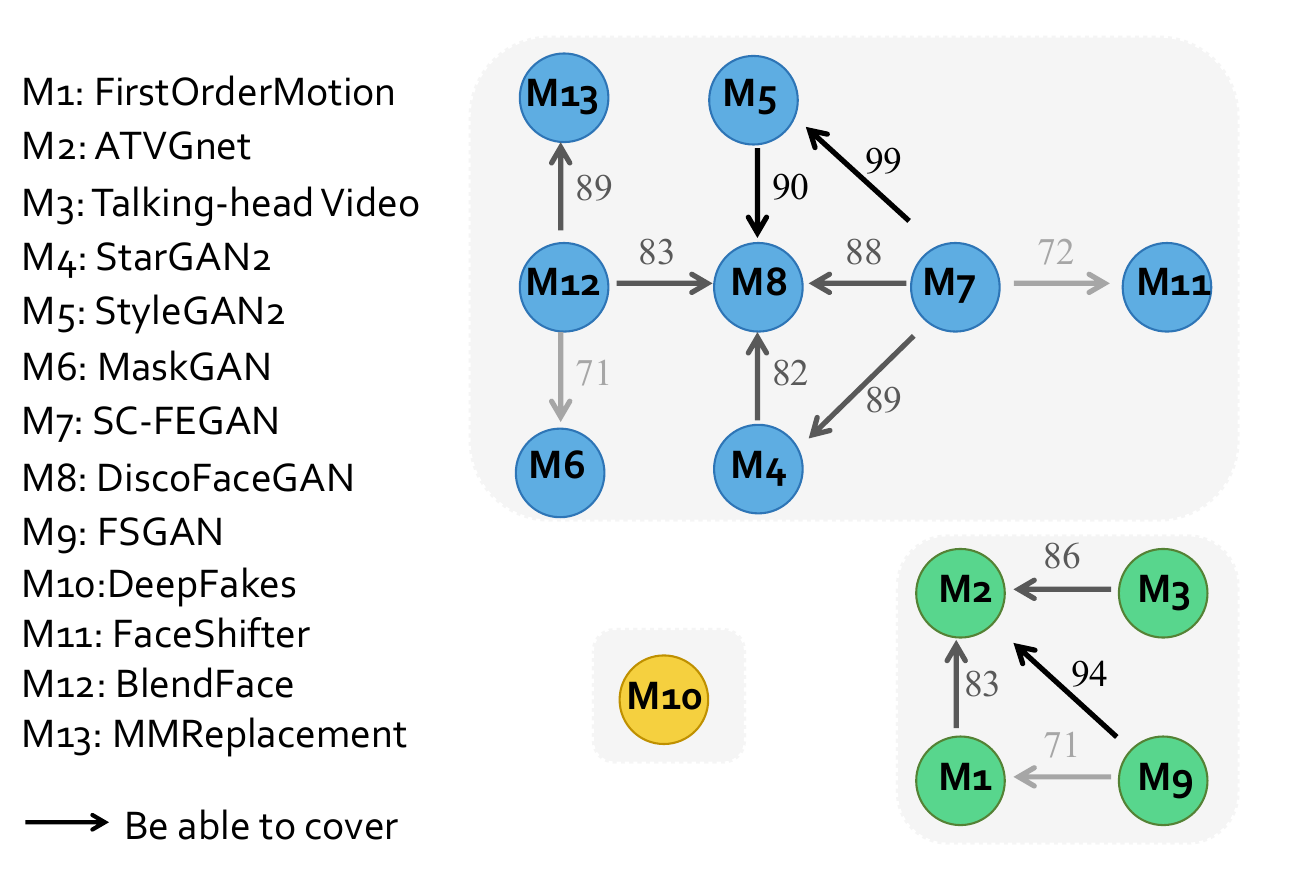}
\end{center}
\cutBeforeFigureCaption
\caption{\textbf{Forgery Taxonomy.} According to pairwise coverage accuracy, forgery approaches are divided into three groups denoted by different colors.
To simplify notations, each forgery approach is assigned a code name.
}
\cutAfterFigureCaption
\label{fig:coverage}
\end{figure}

To better illustrate coverage relationships, we further build a forgery taxonomy in Fig.~\ref{fig:coverage}, which is a directed graph with edges indicating high cross-forgery accuracy. We manually set a threshold of $70\%$, \ie only preserve edges A$\rightarrow$B satisfying that the detection model trained on data of approach A has accuracy no lower than $70\%$ on the test set of approach B. When viewed as an undirected graph, our forgery taxonomy has three connected components. In other words, our thirteen forgery approaches can be divided into three groups, and there is no strong coverage relationship (accuracy $\geq70\%$) between any pair of approaches from different groups.

\noindent\textbf{Dataset Construction.}
We in total set up four few-shot forgery detection datasets within our data base.
For each dataset, $5\sim8$ forgery approaches are picked as majority classes. We then carefully select another forgery approach as the minority class, which should not be well covered by any of the majority classes in Fig.~\ref{fig:heat-cross}.
In the training set, we include all data from these majority forgery classes, and sample an appropriate number of real data ($0.5$M) to balance the distribution of real and fake. For the minority class, we use the $50$-shot setting by default, \ie $50$ samples from the minority class for training.
This number is chosen based on two considerations: \textbf{1)} acquiring $50$ samples is generally feasible even if we do not know full details of the novel forgery approach or the reproduction cost is high, and \textbf{2)} compared to the sheer size ($>20$k) of majority classes, $50$ data are needed to roughly estimate the distribution of the minority class. Otherwise, prediction results on the minority class are hard to be statistically significant and may be tantamount to blind guesses.
For testing, we use all test data of majority classes as well as the minority class. 

\begin{table}[!t]
\centering
\cutBeforeTableCaption
\caption{\textbf{Few-shot Forgery Detection Datasets.}
Each few-shot forgery detection dataset includes $5\sim8$ forgery approaches as majority classes and another one as the minority class. Code names are defined in Fig.~\ref{fig:coverage}.
}
\cutAfterTableCaption
\scalebox{0.8}{
\begin{tabular}{c|c|c}
\hline
Name         & Majority                    & Minority             \\ \hline \hline
Group1\_FSG  & M4,M5,M6,M7,M8,M11,M12,M13  &  M9                  \\ \hline
Group1\_DFL  & M4,M5,M6,M7,M8,M11,M12,M13  &  M10                 \\ \hline
FR+FE\_FSG   & M1,M2,M3,M4,M5,M6,M7,M8     &  M9                  \\ \hline
FR+FS\_SG    & M1,M2,M3,M9,M10,M11         &  M4                  \\ \hline
\end{tabular}
}
\label{tbl:datasets}
\cutAfterTableCaption
\end{table}

We design two different standards to choose the approaches and form our four datasets. An overview of these datasets is given in Tab.~\ref{tbl:datasets}.
The first standard follows three groups divided according to our forgery taxonomy in Fig.~\ref{fig:coverage}. 
We choose the largest group (M4, M5, M6, M7, M8, M11, M12, M13) as majority classes, and select the minority class from the other two groups. We include two choices for the minority class, \ie M9 and M10 in our Group1\_FSG and Group1\_DFL datasets, respectively.
On the other hand, our forgery approaches can also be divided into four categories - face reenactment, face editing, face transfer, and face swap. 
We base our second standard for selecting forgery approaches on this new divide with four categories, since it differs from the previous divide with three groups, 
We select two categories as majority classes, and then pick the minority class from a third category.
The dataset name FR+FE\_FSG refers to Face Reenactment and Face Editing as majority, while M9 from Face Swap as minority.
Similarly, for FR+FS\_SG, majority classes are from Face Reenactment and Face Swap, and we pick M4 from Face Editing as the minority class.

\subsection{Benchmark Evaluation}
For evaluation, we follow the tradition of previous forgery detection and only report metrics of binary classification.
To utilize multi-class labels and extract richer features, the detection method is firstly trained to give multi-class predictions, and then transfer to binary classification by simply mapping $n+1$ forgery classes to one single class ``fake", \ie $\{1,2,\cdots,n+1\}\rightarrow1$.
To fully evaluate the quality of a method, four evaluation metrics are used: \textbf{1)} accuracy score on the minor class ($\text{ACC}_\text{minor}$); \textbf{2)} True Positive Rate (TPR) of the minor class at False Positive Rate (FPR) equal to $0.1\%$ (TPR$^{0.1\%}_\text{minor}$); \textbf{3)} accuracy score on the whole test set ($\text{ACC}_\text{all}$); \textbf{4)} area under the receiver operating characteristic curve computed with the whole test set ($\text{AUC}$).

\cutBeforeSection
\section{Guided Adversarial Interpolation}
\cutAfterSection


\begin{figure*}
\begin{center}
\includegraphics[width=1.0\linewidth]{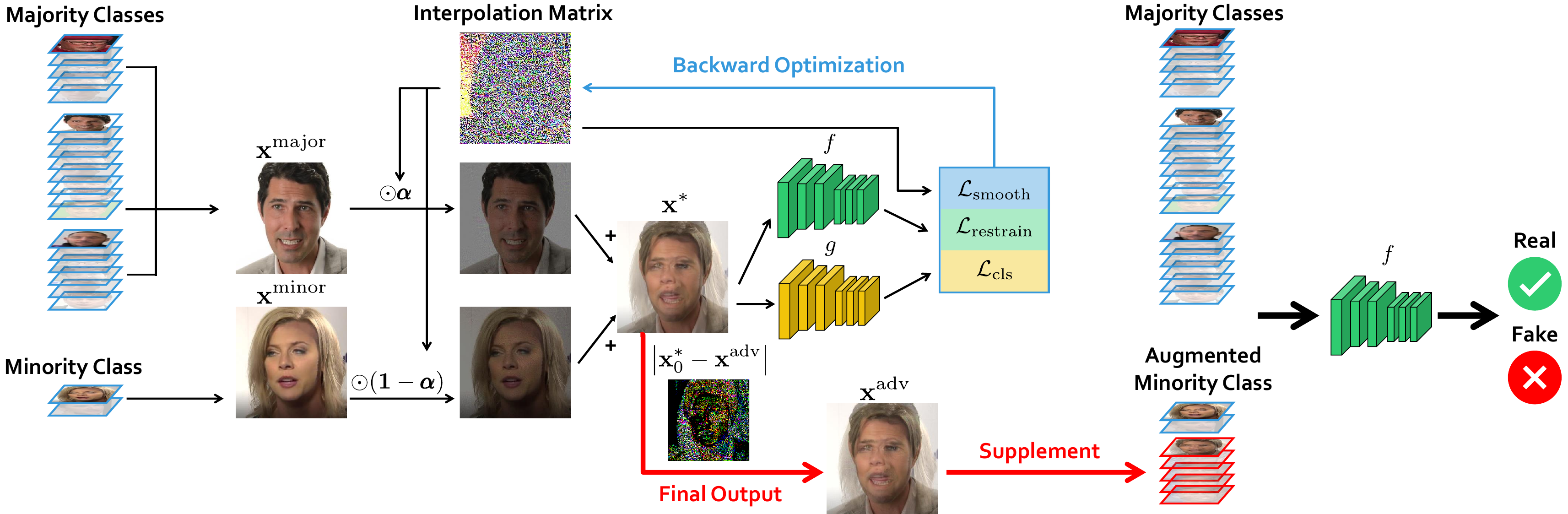}
\end{center}
\cutBeforeFigureCaption
\caption{\textbf{Method Overview.} 
Our GAI interpolates a majority sample and a minority sample under the guidance of a pre-trained teacher network $g$. 
Generated samples $\mathbf{x}^\text{adv}$ after iterative updates are added to the minority class, and help to train the student network $f$ for final prediction. $\mathbf{x}_0^*$ denotes initially generated $\mathbf{x}^*$ with $\pmb{\alpha}$ filled with constant initialization value $\alpha_0$.
}
\cutAfterFigureCaption
\label{fig:pipeline}
\end{figure*}

In this section, we state the motivation for designing our proposed GAI, and give the framework of our method.

\subsection{Framework of GAI}

Collecting a large amount of data from an unknown forgery approach is usually infeasible, since most forgery technologies are not publicly available. Therefore, a key target for few-shot forgery detection is to expand information of the minority domain.
Compared to simple data augmentations within the minority domain, transferring information from majority domains to the minority domain is more reasonable.
Inspired by M2m~\cite{kim2020m2m}, the guidance by a given (teacher) network $g$ during the transferring process helps to generate more informative new samples for the final (student) network $f$.
Therefore, we propose Guided Adversarial Interpolation (GAI) that adversarially interpolates forgery artifacts of minority samples with majority samples under the guidance of a teacher network.
The whole pipeline is shown in Fig.~\ref{fig:pipeline}.

To transfer the information from majority domains to the minority domain, we sample $\mathbf{x}^\text{major}$ from one majority class $c^\text{major}$ and transform it through:
\begin{align}
\mathcal{T}: \mathbf{x}^\text{major} \rightarrow \mathbf{x}^{*},
\end{align}
where $\mathcal{T}$ is a flexible transform function (the choice of $\mathcal{T}$ will be discussed in Sec.~\ref{subsec:at}), and $\mathbf{x}^{*}$ is the transformed sample that is aimed at augmenting the minority class.
The generated samples will be labeled as the minority class $y^* = n+1$ and mixed into $\mathcal{D}^\text{train}$.
The final prediction network $f$ will be trained on this augmented training set.

To guide this transform, we simply train a basic classifier $g$ on $\mathcal{D}^\text{train}$. The $g$ will serve as the teacher network. Under the guidance of $g$, we iteratively update generated $\mathbf{x}^*$ for $T$ steps and finally obtain an optimal $\mathbf{x}^\text{adv}$.
Note that $g$ does not necessarily generalize well to $d^\text{test}_{n+1}$.
We find the basic classifier $g$ is already able to provide guidance for the transformation process.
The objective function for iterative updates is:
\begin{align}
\mathbf{x}^\text{adv} 
&= \arg \min_{\mathbf{x}^*} \mathcal{L}(\mathbf{x}^*), \nonumber \\
\mathcal{L}(\mathbf{x}^*) &= \mathcal{L}_\text{cls}(\mathbf{x}^*; g)+\lambda \cdot \mathcal{L}_\text{restrain}(\mathbf{x}^*; f)+\beta\cdot\mathcal{L}_\text{smooth}.
\label{eq:method-objective}
\end{align}
The first term $\mathcal{L}_\text{cls}(\mathbf{x}^*;g)$ is a common cross-entropy loss which requires the teacher $g$ to classify the generated $\mathbf{x}^*$ into the minority class $n+1$.
The second term $\mathcal{L}_\text{restrain}(\mathbf{x}^{*};f)=f_{c^\text{major}}(\mathbf{x}^{*})$ prevents the student $f$ from predicting $\mathbf{x}^{*}$ to be its original class $c^\text{major}$, and thus enhances training stability.
In addition, the third term $\mathcal{L}_\text{smooth}$ is a smoothness constraint to guarantee perceptual quality after the transform.
Parameters $\lambda$ and $\beta$ are used to control the balance among three terms.
The whole objective function involves both the teacher $g$ and the student $f$ to the iterative process for transformation, utilizing the knowledge extraction capabilities of both networks.

Notably, although forgery detection is a binary classification task, $g$ is designed as a multi-class classification network. 
Otherwise, $g$ will not be able to tell apart the minority domain from other majority forgery domains.

\subsection{Adversarial Interpolation}
\label{subsec:at}

\begin{algorithm}[!t]
    \centering
    \scalebox{0.9}{
    \begin{minipage}{0.98\linewidth}
    \DontPrintSemicolon
    \SetAlgoLined
    \SetKwInput{KwInput}{Input}
    \SetKwInput{KwOutput}{Output}
    \KwInput{Majority samples $\mathcal{B}=\left\{\left(\mathbf{x}_i,y_i\right)\right\}_{i=1}^b$. Minority samples $\left\{\left(\mathbf{x}_{b+i},n+1\right)\right\}_{i=1}^k$. A student model $f$. A teacher model $g$. Number of steps $T$, step size $\eta$, $\lambda$, $\beta>0$. Replacement probability $p$, initialization value $\alpha_0$, rejection threshold $\tau\in[0,1]$.}
    \KwOutput{Updated few-shot samples $\left\{\mathbf{x}'_{b+i}\right\}_{i=1}^k$}
    \For{$i\gets1$ \KwTo $k$}{
        $\mathbf{x}'_{b+i}\gets\mathbf{x}_{b+i}$\;
        $z\sim\text{Bernoulli}(p)$\;
        \If{$z=1$}{
            $\pmb{\alpha}\gets\alpha_0\cdot\mathbf{1}^{H\times W\times D}$\;
            Randomly perturb $\pmb{\alpha}$ with a small noise\;
            $\mathbf{x}^\text{major}\gets$ A random $\mathbf{x}_{r}$ in $\mathcal{B}$ ($1\leq r\leq b$)\;
            $\mathbf{x}^\text{minor}\gets\mathbf{x}_{b+i}$\;
            \For{$j\gets1$ \KwTo $T$}{
                $\mathbf{x}^*\gets\pmb{\alpha}\odot\mathbf{x}^\text{major}+\left(1-\pmb{\alpha}\right)\odot\mathbf{x}^\text{minor}$\;
                $\pmb{\xi}\gets\nabla_{\pmb{\alpha}}\mathcal{L}(\mathbf{x}^*)$\;
                $\pmb{\alpha}\gets\pmb{\alpha}-\eta\cdot\pmb{\xi}/\left\Vert\pmb{\xi}\right\Vert_2$\;
                $\pmb{\alpha}\gets\max\left\{\min\left\{\pmb{\alpha},\mathbf{1}\right\},\mathbf{0}\right\}$\;
            }
            $\mathbf{x}^\text{adv}\gets\pmb{\alpha}\odot\mathbf{x}^\text{major}+\left(1-\pmb{\alpha}\right)\odot\mathbf{x}^\text{minor}$\;
            \If{$\textnormal{softmax}\left(g(\mathbf{x}^\textnormal{adv})\right)_{n+1}\geq\tau$}{
                $\mathbf{x}'_{b+i}\gets\mathbf{x}^\text{adv}$\;
            }
        }
    }
    \caption{Guided Adversarial Interpolation}
    \cutAfterFigureCaption
    \label{alg:gai}
    \end{minipage}%
}
\end{algorithm}

\begin{figure}[!t]
\begin{center}
\includegraphics[width=0.9\linewidth]{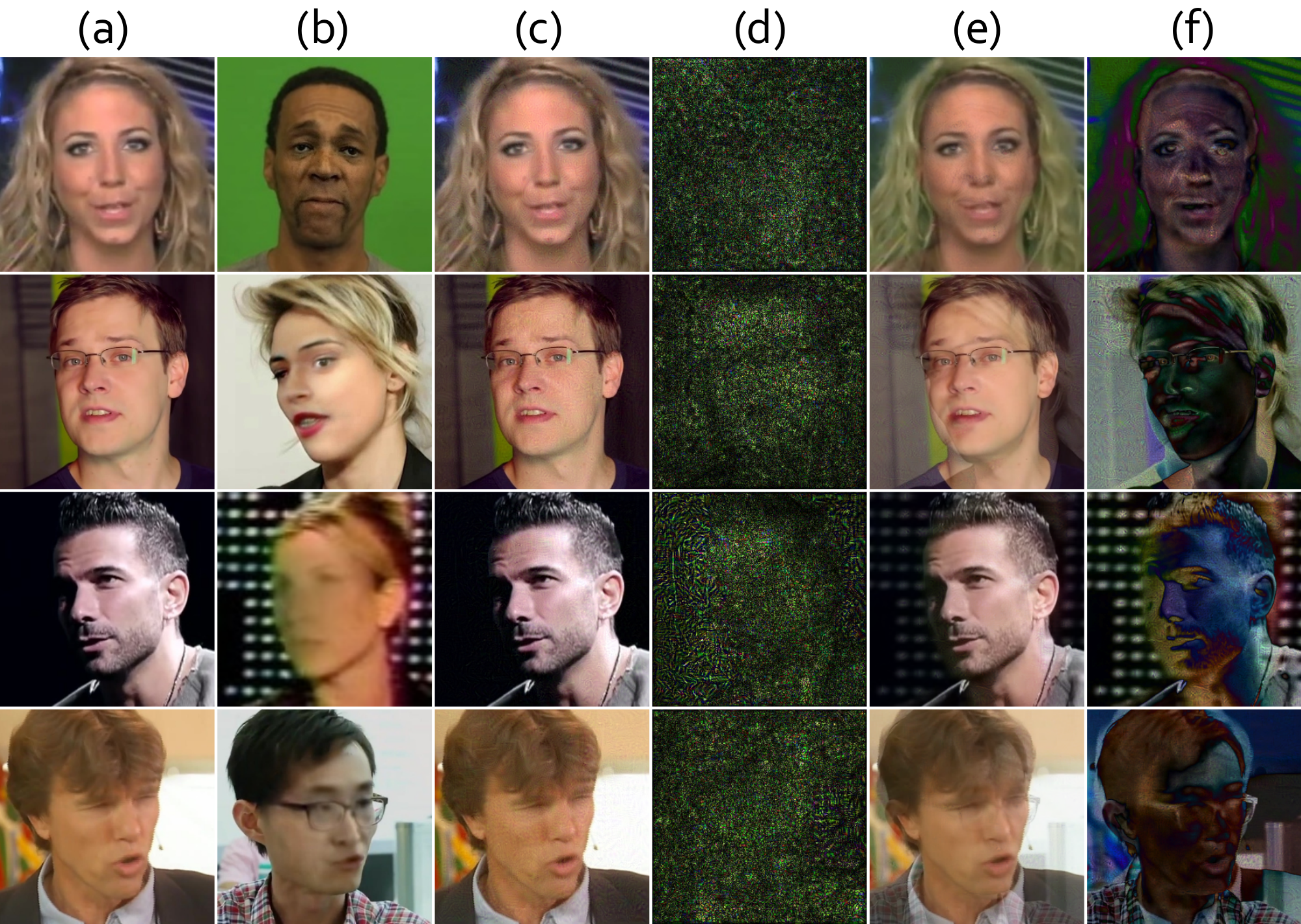}
\end{center}
\cutBeforeFigureCaption
\caption{\textbf{Visualization of Transformed Samples by GAI.} (a) Majority samples. (b) Minority samples. (c) Samples generated by GAI$-$. (d) Difference between (a) and (c) magnified by $15\times$. (e) Samples generated by GAI. (f) Difference between (a) and (e) magnified by $3\times$. Compared to disordered perturbations added by GAI$-$, GAI attaches visually discernable and cognitively meaningful patterns to majority samples. 
}
\cutAfterFigureCaption
\label{fig:vis}
\end{figure}

For the choice of the transform function $\mathcal{T}$, existing work ~\cite{kim2020m2m} obtains the transformed image $\mathbf{x}^{*}$ by adding a perturbation directly:
\begin{align}
\mathbf{x}^{*} = \mathbf{x}^\text{major} + \pmb{\delta}.
\label{eq:method-transform}
\end{align}
We denote the method using this perturbation-based transform as GAI$-$. This transform introduces two problems in this task: \textbf{1)} similar to adversarial examples~\cite{goodfellow2015explaining}, generated $\mathbf{x}^\text{adv}$ after iterations is usually still very close to $\mathbf{x}^\text{major}$,
which is verified in Fig.~\ref{fig:vis}; \textbf{2)} generated samples only contain information of majority domains but lack that of the minority domain.
To address these two problems, we use Adversarial Interpolation~\cite{qiu2020semanticadv} as our transform function $\mathcal{T}$:
\begin{align}
\mathbf{x}^{*} &= \pmb{\alpha} \odot \mathbf{x}^\text{major}
+ (\mathbf{1} - \pmb{\alpha}) \odot \mathbf{x}^\text{minor},
\label{eq:method-interp}
\end{align}
where ${\pmb{\alpha}} \in \mathbb{R}^{H \times W \times D}$ ($\alpha_{h,w,d} \in [0, 1]$) is an interpolation coefficient tensor with the same shape as the input image.
Unlike Equation~\ref{eq:method-transform} which only needs a majority sample $\mathbf{x}^\text{major}$ during the transform, it requires an additional minority sample $\mathbf{x}^\text{minor}$, and embeds more forgery traces of minority samples to augmented ones via optimizing $\pmb{\alpha}$. Objective function~\ref{eq:method-objective} will adversarially move the interpolated sample to be labeled as the minority class.
In addition, through Equation~\ref{eq:method-interp}, the problem that $\mathbf{x}^\text{adv}$ is overly close to $\mathbf{x}^\text{major}$ is also simply resolved by giving $\pmb{\alpha}$ a moderate initial value $\alpha_0\in(0,1)$. 
The detailed procedure of GAI is presented in Algorithm~\ref{alg:gai}.

To encourage the interpolation coefficient tensor $\pmb{\alpha}$ to be locally smooth, we apply total variation loss to $\pmb{\alpha}$, which has been widely used as a pixel-wise de-noising objective for natural image processing~\cite{mahendran2015understanding}.
Thereupon, the smoothness constraint $\mathcal{L}_\text{smooth}$ is designed as:
\begin{equation}
\begin{split}
\mathcal{L}_\text{smooth}(\pmb{\alpha}) =& \frac{1}{(H-1)W}\sum_{h=1}^{H-1} \sum_{w=1}^{W} \| {\pmb{\alpha}}_{h+1,w} - {\pmb{\alpha}}_{h,w}\|^2_2  \\   
& + \frac{1}{H(W-1)}\sum_{h=1}^{H} \sum_{w=1}^{W-1} \|{\pmb{\alpha}}_{h,w+1} - {\pmb{\alpha}}_{h,w}\|^2_2.
\end{split}
\end{equation}

We select some data from our four datasets, and visualize samples generated by our proposed GAI as well as the degenerated GAI$-$ without adversarial interpolation. As shown in Fig.~\ref{fig:vis}, generated images from GAI$-$ are almost the same as the corresponding majority samples, and do not give out any visually effective information about the minority class. On the other hand, data generated by GAI combine both majority and minority samples. Ablation experiments in Sec.~\ref{subsec:exp-ablation} demonstrate that our method performs better when generated $\mathbf{x}^\text{adv}$ has visible differences with $\mathbf{x}^\text{major}$.

\subsection{Detailed Components of GAI}

\noindent\textbf{Factors of Adversarial Interpolation.}
Our proposed GAI has various degraded versions when some factors are modified or removed.
As an example, if the interpolation coefficient tensor $\pmb{\alpha}$ is filled with a constant scalar, and the generated sample $\mathbf{x}^{*}$ is assigned with a correspondingly interpolated label, our method will degenerate to mixup~\cite{zhang2017mixup} with a fixed mixing ratio.

Although mixup easily gains accuracy improvement in many general classification tasks, the interpolated label is not an optimal choice in the few-shot forgery detection scenario. We also study the case of interpolation without teacher guidance to check the contribution of the teacher network.
Relevant discussions are provided in Sec.~\ref{subsec:exp-ablation}.

\noindent\textbf{Sample Replacement Strategy.}
In practical implementation, we first upsample the minority data for class balancing. We then replace the repeated few-shot samples with the ones generated by GAI with a probability of $p$. Moreover, to have stronger guidance from the teacher, we enforce that $g$ should classify generated samples into the few-shot class. Specifically, we reject a generated sample $\mathbf{x}^\text{adv}$ if its confidence on the few-shot class predicted by $g$, \ie $\text{softmax}\left(g(\mathbf{x}^\text{adv})\right)_{n+1}$, is lower than a threshold $\tau$.


\cutBeforeSection
\section{Experiments}
\cutAfterSection
\label{sec:exp}

\subsection{Experimental Settings}

We compare our proposed GAI with the following representative methods on the few-shot forgery detection task.

\noindent\textbf{Unseen} only uses $\mathcal{D}^{\text{train}}_{\text{base}}$ for training, investigating the coverage of majority classes on the minority forgery approach.

\noindent\textbf{Instance-Balanced Sampling (IB)} is the most common way of sampling data for balanced datasets, where each sample has an equal probability to be selected during training.

\noindent\textbf{Class-Balanced Sampling (CB)} is the most common way of sampling data for imbalanced datasets, where each class has an equal probability to be selected during training.

\noindent\textbf{ForensicTransfer (FT)}~\cite{cozzolino2018forensictransfer} uses an auto-encoder based structure to disentangle forged and real samples in the latent space. It predicts the class by comparing the activated feature maps.

\noindent\textbf{QDL-CMFD (QDL)}~\cite{ARIA2022213} is a method to detect copy-move through manipulation detection and similarity detection. We adopt its framework manipulation detection for general forgery detection.

\noindent\re{\textbf{SBI}~\cite{shiohara2022detecting} blends pseudo source and target images from single pristine images as fake samples for training. Here we use the official pre-trained SBI model.}

\noindent\textbf{One-Shot Face Generation (OSFG)}~\cite{yang2020one} fine-tunes a pre-trained image generator with one minority sample. For convenience, we switch its original generator to StyleGAN2~\cite{Karras2019stylegan2} in our implementation.
These samples are added as new minority samples to expand the information of the minority class.

\noindent\textbf{GAI$-$} mentioned in Sec.~\ref{subsec:at} is also included for comparison. Instead of involving the interpolation coefficient tensor $\pmb{\alpha}$, GAI$-$ only uses the perturbation-based transform, serving as an ablation study.

\subsection{Implementation Details.}
We select Xception~\cite{chollet2017xception} as the backbone network for all experiments, since it has been widely used by various forgery detection methods. For each few-shot forgery detection dataset, we first train a base model on the base dataset $\mathcal{D}^{\text{train}}_{\text{base}}$ with only majority classes, which is also the ``unseen" baseline. Then, we start from the base model for further experiments on subsequent few-shot forgery detection tasks.

For training the CB baseline, we duplicate the $50$ minority training samples by $2$k times \re{to achieve class-balanced sampling}. We directly use the CB baseline as the teacher model for our proposed GAI. For hyper-parameters in GAI, we by default set $T=10$, $\lambda=\tau=0.5$, $\beta=10$, $\eta=1$, $p=0.99$, and $\alpha_0=0.75$.

And for training the base model, \ie the ``unseen" baseline, we start with an ImageNet pre-trained checkpoint, and follow the optimization schedule with $100$k iterations as described in \cite{he2021forgerynet}. We set the learning rate to $0.05$ and weight decay to $10^{-4}$.

We start from the base model for further experiments on few-shot forgery detection methods. Specifically, we initialize the backbone network with weights of the base model when training our proposed method and other baselines. For these experiments starting from the base model, we use SGD with Nesterov momentum $0.9$ as the optimizer and weight decay $10^{-4}$, yet we adopt a much shorter schedule of 30k iterations and a $10\times$ smaller learning rate ($0.005$). We perform a linear warm-up on the learning rate during the first 1k iterations, and decay it with a factor of $0.1$ at $10$k and $20$k. We run each experiment three times, and report the mean value and standard deviation of metrics.


\cutBeforeTableCaption
\begin{table*}[!t]
\centering
\caption{\textbf{Comparison Results on Established Benchmark.} Compared to other state-of-the-art methods, our GAI achieves the best performances, and is robust to choices of majority and minority forgery approaches. }
\cutAfterTableCaption
\begin{subtable}[h]{0.9\textwidth}
\centering
\cutBeforeSubTableCaption
\caption{Benchmarking results on Group1\_FSG and Group1\_DFL.}
\cutAfterSubTableCaption
\scalebox{0.8}{
\begin{tabular}{c|c|c|c|c|c|c|c|c}
\hline
Dataset & \multicolumn{4}{c}{Group1\_FSG} \vline & \multicolumn{4}{c}{Group1\_DFL} \\ \hline
Metric & $\text{ACC}_{\text{minor}}$      & TPR$^{0.1\%}_{\text{minor}}$       & $\text{ACC}_{\text{all}}$      & AUC      & $\text{ACC}_{\text{minor}}$      & TPR$^{0.1\%}_{\text{minor}}$       & $\text{ACC}_{\text{all}}$      & AUC     \\ \hline \hline
Unseen & $26.61_{\pm3.32}$ & $30.01_{\pm2.48}$ & $94.79_{\pm0.24}$ & $98.33_{\pm0.45}$ & $46.10_{\pm10.85}$ & $48.80_{\pm10.14}$ & $95.91_{\pm0.79}$ & $98.89_{\pm0.38}$          \\ \hline
IB & $35.23_{\pm1.54}$ & $38.85_{\pm2.65}$ & $95.37_{\pm0.10}$ & $99.56_{\pm0.03}$ & $55.86_{\pm2.67}$ & $64.73_{\pm0.61}$ & $96.61_{\pm0.19}$ & $99.87_{\pm0.01}$          \\ \hline
CB & $75.14_{\pm1.07}$ & $75.72_{\pm1.32}$ & $98.07_{\pm0.08}$ & $99.66_{\pm0.02}$ & $91.61_{\pm0.88}$ & $92.28_{\pm0.60}$ & $99.15_{\pm0.06}$ & $99.93_{\pm0.03}$          \\ \hline
FT & $43.99_{\pm2.88}$ & - & $86.95_{\pm1.64}$ & - & $88.98_{\pm2.81}$ & - & $91.84_{\pm0.92}$ & -          \\ \hline
QDL & $64.40_{\pm0.87}$ & $96.25_{\pm3.21}$ & $90.82_{\pm0.18}$ & $96.90_{\pm0.30}$ & $32.82_{\pm0.80}$ & $\textbf{98.78}_{\pm0.57}$ & $88.41_{\pm0.19}$ & $95.85_{\pm0.14}$          \\ \hline
\re{SBI} & \re{$58.50$} & \re{$\textbf{98.11}$} & \re{$61.69$} & \re{$67.66$} & \re{$70.22$} & \re{$93.97$} & \re{$62.56$} & \re{$68.83$}          \\ \hline
OSFG & $76.57_{\pm1.20}$ & $77.47_{\pm1.18}$ & $98.18_{\pm0.09}$ & $99.75_{\pm0.01}$ & $92.62_{\pm1.34}$ & $93.88_{\pm1.11}$ & $99.23_{\pm0.09}$ & $99.95_{\pm0.00}$          \\ \hline
GAI$-$ & $74.92_{\pm1.22}$ & $75.96_{\pm1.46}$ & $98.07_{\pm0.09}$ & $99.75_{\pm0.03}$ & $93.17_{\pm0.30}$ & $94.00_{\pm0.32}$ & $99.28_{\pm0.02}$ & $99.95_{\pm0.00}$          \\ \hline
GAI & $\textbf{78.89}_{\pm0.63}$ & $79.07_{\pm0.65}$ & $\textbf{98.34}_{\pm0.04}$ & $\textbf{99.77}_{\pm0.03}$ & $\textbf{95.15}_{\pm0.30}$ & $95.91_{\pm0.40}$ & $\textbf{99.41}_{\pm0.02}$ & $\textbf{99.96}_{\pm0.00}$          \\ \hline
\end{tabular}
}
\label{tbl:benchmark_sub1}
\end{subtable}
\begin{subtable}[h]{0.9\textwidth}
\centering
\cutBeforeSubTableCaption
\caption{Benchmarking results on FR+FE\_FSG and FR+FS\_SG.}
\cutAfterSubTableCaption
\scalebox{0.8}{
\begin{tabular}{c|c|c|c|c|c|c|c|c}
\hline
Dataset & \multicolumn{4}{c}{FR+FE\_FSG} \vline & \multicolumn{4}{c}{FR+FS\_SG} \\ \hline
Metric & $\text{ACC}_{\text{minor}}$      & TPR$^{0.1\%}_{\text{minor}}$       & $\text{ACC}_{\text{all}}$      & AUC      & $\text{ACC}_{\text{minor}}$      & TPR$^{0.1\%}_{\text{minor}}$       & $\text{ACC}_{\text{all}}$      & AUC     \\ \hline \hline
Unseen & $59.98_{\pm2.25}$ & $43.58_{\pm5.83}$ & $96.62_{\pm0.10}$ & $99.53_{\pm0.04}$ & $39.90_{\pm1.67}$ & $19.01_{\pm0.78}$ & $96.54_{\pm0.11}$ & $99.15_{\pm0.02}$         \\ \hline
IB & $59.84_{\pm3.36}$ & $41.03_{\pm2.09}$ & $96.64_{\pm0.22}$ & $99.68_{\pm0.01}$ & $48.95_{\pm2.08}$ & $23.95_{\pm1.33}$ & $96.98_{\pm0.09}$ & $99.52_{\pm0.06}$          \\ \hline
CB & $84.93_{\pm1.07}$ & $74.42_{\pm2.20}$ & $98.35_{\pm0.07}$ & $99.81_{\pm0.00}$ & $71.07_{\pm0.65}$ & $59.48_{\pm0.95}$ & $98.00_{\pm0.03}$ & $99.57_{\pm0.02}$          \\ \hline
FT & $76.04_{\pm1.01}$ & - & $89.61_{\pm0.23}$ & - & $\textbf{87.21}_{\pm2.76}$ & - & $91.07_{\pm0.30}$ & -          \\ \hline
QDL & $60.69_{\pm0.91}$ & $91.87_{\pm3.68}$ & $89.74_{\pm0.19}$ & $96.94_{\pm0.29}$ & $54.04_{\pm1.21}$ & $95.87_{\pm4.09}$ & $89.02_{\pm1.22}$ & $96.70_{\pm0.25}$         \\ \hline
\re{SBI} & \re{$58.50$} & \re{$\textbf{98.11}$} & \re{$61.76$} & \re{$67.47$} & \re{$59.17$} & \re{$\textbf{97.27}$} & \re{$61.56$} & \re{$67.25$}          \\ \hline
OSFG & $82.18_{\pm0.87}$ & $71.65_{\pm0.26}$ & $98.13_{\pm0.07}$ & $99.77_{\pm0.03}$ & $70.03_{\pm1.73}$ & $54.34_{\pm2.63}$ & $97.96_{\pm0.07}$ & $99.62_{\pm0.04}$          \\ \hline
GAI$-$ & $84.60_{\pm0.71}$ & $74.39_{\pm1.35}$ & $98.28_{\pm0.06}$ & $99.80_{\pm0.01}$ & $71.29_{\pm1.45}$ & $58.94_{\pm2.63}$ & $98.03_{\pm0.08}$ & $99.61_{\pm0.02}$          \\ \hline
GAI & $\textbf{86.59}_{\pm0.83}$ & $77.01_{\pm1.82}$ & $\textbf{98.44}_{\pm0.04}$ & $\textbf{99.81}_{\pm0.02}$ & $75.17_{\pm0.49}$ & $61.78_{\pm0.42}$ & $\textbf{98.18}_{\pm0.04}$ & $\textbf{99.67}_{\pm0.03}$          \\ \hline
\end{tabular}
}
\label{tbl:benchmark_sub2}
\end{subtable}
\label{tbl:benchmark}
\cutAfterTableCaption
\end{table*}

\subsection{Benchmarking Results}

As presented in Tab.~\ref{tbl:benchmark}, $\text{ACC}_{\text{minor}}$ is universally poor in all datasets for unseen cases, which confirms that picked majority forgery approaches are not able to cover the minority forgery approach.
IB only gains limited improvement on $\text{ACC}_{\text{minor}}$ and $\text{ACC}_{\text{all}}$ due to the low frequency of minority samples.
CB balances class distribution, thus an obvious improvement can be observed in $\text{ACC}_{\text{minor}}$.
However, over-sampling of the minority class limits its generalization.

Our method GAI transfers information from various majority domains to the minority domain by guided adversarial interpolation, and it consistently achieves the best performances on all four datasets.
However, once the key component of adversarial interpolation is omitted (GAI$-$), its performances suffer significant drops.

For the other two more advanced forgery detection methods, FT and OSFG, their performances will fluctuate according to the choices of majority and minority forgery approaches.
FT predicts the class by comparing the activated feature maps in latent space, which introduces a large instability.
Although it owns the highest $\text{ACC}_{\text{minor}}$ on FR+FS\_SG, its $\text{ACC}_{\text{all}}$ drops a lot.
Moreover, all metrics of FT are the lowest in the other three datasets.
For OSFG, it performs well on Group1\_FSG and Group1\_DFL, with all metrics better than CB. Nonetheless, most of its metrics are worse than CB in FR+FE\_FSG and FR+FS\_SG.
QDL is a copy-move detection method and is adopted for this general forgery detection task and gains higher TPR$^{0.1\%}_{\text{minor}}$. \re{More recent SOTA method SBI achieves the highest TPR$^{0.1\%}_{\text{minor}}$ which means that it is safe to trust the SBI when it predicts a sample is real. However, all other metrics of QDL and SBI are significantly worse. It exhibits that SBI and QDL are easy to predict a sample is forged even for a real image.}
Our method GAI achieves the highest scores on most metrics and ranks second on the remaining few. While extreme decision boundaries of some other methods may enhance the performance of a model with respect to individual metrics, our method provides superior performance on all metrics. This indicates that our approach makes a balance prediction for real and forged samples. This superiority is robust to choices of majority and minority forgery approaches, reflecting the robustness of our proposed method.

In addition to comparison with SOTAs, we also have an interesting finding by comparing the $\text{ACC}_{\text{minor}}$ between Group1\_FSG and FR+FE\_FSG.
That is, although they share FSGAN as the minority approach, $\text{ACC}_{\text{minor}}$ on FR+FE\_FSG is significantly higher than that on Group1\_FSG in every method, especially when FSGAN is unseen ($26.61\%$ $\to$ $59.98\%$).
This observation implies that our forgery taxonomy divides forgery approaches better than forgery categories in terms of a cross-forgery relationship.


\subsection{Ablation Study}
\label{subsec:exp-ablation}

\begin{figure}[t]
\begin{center}
\includegraphics[width=0.9\linewidth]{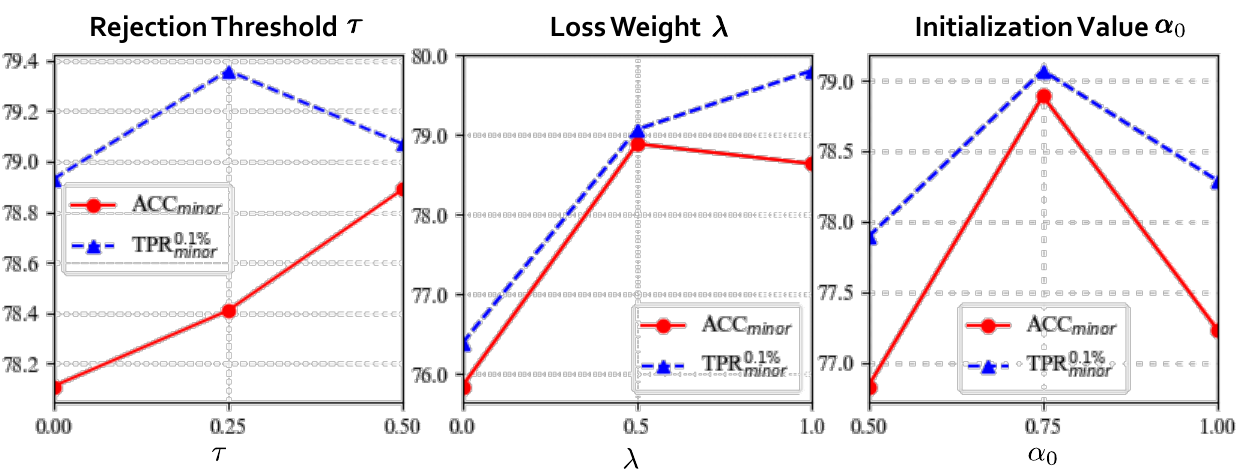}
\end{center}
\cutBeforeFigureCaption
\caption{\textbf{Ablation Experiments on Hyper-parameters.} Setting $\lambda=0$, \ie removing the loss term $\mathcal{L}_{\text{restrain}}(\mathbf{x}^*;f)$ significantly degrades evaluation results. Choosing a proper initialization value $\alpha_0$ for the coefficient $\pmb{\alpha}$ is also important, since higher ($\alpha_0=1$) and lower ($\alpha_0=0.5$) values both cause considerable decreases in performances.
}
\cutAfterFigureCaption
\label{fig:ablation}
\end{figure}

We conduct ablation experiments to study the effects of several hyper-parameters in our proposed GAI. We show results in Tab.~\ref{tbl:ablation} as well as Fig.~\ref{fig:ablation}, and provide detailed analyses in the following paragraphs.

\begin{table}[!t]
\centering
\cutBeforeTableCaption
\caption{\textbf{Ablation Experiments on Teacher Guidance.} Removing the teacher drops performances substantially. Interpolating labels (mixup $0.75$) further lowers the numbers.}
\cutAfterTableCaption
\scalebox{0.8}{
\begin{tabular}{l|c|c|c|c}
\hline
Dataset & \multicolumn{4}{c}{Group1\_FSG} \\ \hline
Metric & $\text{ACC}_{\text{minor}}$      & TPR$^{0.1\%}_{\text{minor}}$       & $\text{ACC}_{\text{all}}$      & AUC \\ \hline \hline
GAI & $\textbf{78.89}_{\pm0.63}$ & $\textbf{79.07}_{\pm0.65}$ & $\textbf{98.34}_{\pm0.04}$ & $\textbf{99.77}_{\pm0.03}$ \\ \hline 
- no teacher & $75.88_{\pm1.42}$ & $75.80_{\pm1.86}$ & $98.10_{\pm0.09}$ & $99.76_{\pm0.00}$ \\ 
- mixup $0.75$ & $72.19_{\pm1.33}$ & $74.50_{\pm1.24}$ & $97.87_{\pm0.09}$ & $99.73_{\pm0.01}$ \\ \hline
\end{tabular}
}
\label{tbl:ablation}
\cutAfterTableCaption
\end{table}

\noindent\textbf{Teacher Guidance.} To validate the importance of the teacher network $g$, we try a much simpler variant of our method where we directly set $\pmb{\alpha}$ to be a constant tensor filled with $\alpha_0=0.75$, and do not perform any further iterative update. This variant entirely removes the dependency on the teacher $g$, and thus we name it ``no teacher". The ``no teacher" variant is also similar to the mixup~\cite{zhang2017mixup}, except that classification labels are not mixed. Therefore, we add another variant ``mixup $0.75$" where we mix both images and labels with the same coefficient $\alpha_0=0.75$. In Tab.~\ref{tbl:ablation}, we observe that these two variants have significantly lower results than our default setting, and mixing labels results in worse performances. These numbers demonstrate that the teacher network $g$ is an indispensable part of our method.

\noindent\textbf{Rejection Threshold $\pmb{\tau}$.} We by default reject GAI-generated samples if the teacher $g$ does not classify it to the minority class with confidence $\geq\tau=0.5$. In this set of experiments, we loosen this restriction to $\tau=0.25$ or even $0$ and see how our method would perform. As shown in Fig.~\ref{fig:ablation}, $\text{ACC}_{\text{minor}}$ gradually decreases within a small range as the threshold $\tau$ becomes lower. Nonetheless, our method reaches a slightly higher TPR$^{0.1\%}_{\text{minor}}$ with $\tau=0.25$.

\noindent\textbf{Loss Weight $\pmb{\lambda}$.} This hyper-parameter $\lambda$ is the weight of the loss term $\mathcal{L}_{\text{restrain}}(\mathbf{x}^*;f)$. We can see from Fig.~\ref{fig:ablation} that removing this loss term, \ie setting $\lambda=0$, results in significantly worse performances than our default setting. On the other hand, increasing $\lambda$ to $1$ gives results similar to $\lambda=0.5$, with TPR$^{0.1\%}_{\text{minor}}$ even higher.

\noindent\textbf{Initialization Value $\pmb{\alpha}_{\mathbf{0}}$.} The initialization for the coefficient tensor $\pmb{\alpha}$ is also important. We find that initializing with $\alpha_0=0.5$ or $1$ leads to substantial performance drops compared to the default $\alpha_0=0.75$. Note that setting $\alpha_0=1$ initializes $\mathbf{x}^*$ at the majority sample, which is the same as the GAI$-$ baseline and would cause final $\mathbf{x}^{\text{adv}}$ to be visually similar to $\mathbf{x}^{\text{major}}$. Nevertheless, results of $\alpha_0=1$ are better than those of GAI$-$.

\subsection{Number of Shots}

\begin{table*}[!t]
\centering
\cutBeforeTableCaption
\caption{\textbf{Comparison Results on Established $10$-shot Benchmark.} Compared to other state-of-the-art methods, our GAI achieves the best performances, and is robust to choices of majority and minority forgery approaches. }
\cutAfterTableCaption
\begin{subtable}[h]{0.9\textwidth}
\centering
\cutBeforeSubTableCaption
\caption{$10$-shot Benchmarking results on Group1\_FSG and Group1\_DFL.}
\cutAfterSubTableCaption
\scalebox{0.8}{
\begin{tabular}{c|c|c|c|c|c|c|c|c}
\hline
Dataset & \multicolumn{4}{c}{Group1\_FSG} \vline & \multicolumn{4}{c}{Group1\_DFL} \\ \hline
Metric & $\text{ACC}_{\text{minor}}$      & TPR$^{0.1\%}_{\text{minor}}$       & $\text{ACC}_{\text{all}}$      & AUC      & $\text{ACC}_{\text{minor}}$      & TPR$^{0.1\%}_{\text{minor}}$       & $\text{ACC}_{\text{all}}$      & AUC     \\ \hline \hline
Unseen & $26.61_{\pm3.32}$ & $30.01_{\pm2.48}$ & $94.79_{\pm0.24}$ & $98.33_{\pm0.45}$ & $46.10_{\pm10.85}$ & $48.80_{\pm10.14}$ & $95.91_{\pm0.79}$ & $98.89_{\pm0.38}$          \\ \hline
IB & $27.97_{\pm1.72}$ & $31.27_{\pm3.25}$ & $94.89_{\pm0.12}$ & $99.03_{\pm0.02}$ & $54.34_{\pm5.82}$ & $55.96_{\pm5.14}$ & $96.52_{\pm0.41}$ & $99.49_{\pm0.08}$          \\ \hline
CB & $39.61_{\pm3.38}$ & $43.86_{\pm1.97}$ & $95.67_{\pm0.22}$ & $99.18_{\pm0.11}$ & $78.66_{\pm2.68}$ & $80.59_{\pm2.05}$ & $98.23_{\pm0.19}$ & $99.83_{\pm0.01}$          \\ \hline
FT & $30.81_{\pm2.72}$ & - & $86.11_{\pm1.27}$ & - & $77.81_{\pm0.67}$ & - & $90.48_{\pm0.69}$ & -          \\ \hline
QDL & $\textbf{64.21}_{\pm1.03}$ & $\textbf{96.94}_{\pm2.00}$ & $90.91_{\pm0.04}$ & $97.07_{\pm0.25}$ & $32.81_{\pm1.92}$ & $\textbf{99.57}_{\pm0.13}$ & $88.60_{0.14}$ & $95.62_{\pm0.20}$          \\ \hline
OSFG & $47.97_{\pm0.23}$ & $49.46_{\pm1.60}$ & $96.24_{\pm0.00}$ & $99.47_{\pm0.01}$ & $79.22_{\pm3.73}$ & $81.75_{\pm3.22}$ & $98.28_{\pm0.25}$ & $99.89_{\pm0.01}$          \\ \hline
GAI$-$ & $45.63_{\pm2.38}$ & $48.51_{\pm1.29}$ & $96.09_{\pm0.17}$ & $\textbf{99.50}_{\pm0.05}$ & $81.38_{\pm0.70}$ & $83.73_{\pm0.64}$ & $98.43_{\pm0.06}$ & $99.89_{\pm0.01}$          \\ \hline
GAI & $52.33_{\pm0.76}$ & $55.07_{\pm1.98}$ & $\textbf{96.53}_{\pm0.05}$ & $99.49_{\pm0.04}$ & $\textbf{88.38}_{\pm0.64}$ & $88.77_{\pm1.08}$ & $\textbf{98.93}_{\pm0.05}$ & $\textbf{99.93}_{\pm0.00}$          \\ \hline
\end{tabular}
}
\label{tbl:benchmark_sub1_10shot}
\end{subtable}
\begin{subtable}[h]{0.9\textwidth}
\centering
\cutBeforeSubTableCaption
\caption{$10$-shot Benchmarking results on FR+FE\_FSG and FR+FS\_SG.}
\cutAfterSubTableCaption
\scalebox{0.8}{
\begin{tabular}{c|c|c|c|c|c|c|c|c}
\hline
Dataset & \multicolumn{4}{c}{FR+FE\_FSG} \vline & \multicolumn{4}{c}{FR+FS\_SG} \\ \hline
Metric & $\text{ACC}_{\text{minor}}$      & TPR$^{0.1\%}_{\text{minor}}$       & $\text{ACC}_{\text{all}}$      & AUC      & $\text{ACC}_{\text{minor}}$      & TPR$^{0.1\%}_{\text{minor}}$       & $\text{ACC}_{\text{all}}$      & AUC     \\ \hline \hline
Unseen & $59.98_{\pm2.25}$ & $43.58_{\pm5.83}$ & $96.62_{\pm0.10}$ & $99.53_{\pm0.04}$ & $39.90_{\pm1.67}$ & $19.01_{\pm0.78}$ & $96.54_{\pm0.11}$ & $99.15_{\pm0.02}$         \\ \hline
IB & $61.21_{\pm2.55}$ & $42.87_{\pm1.94}$ & $96.74_{\pm0.16}$ & $99.61_{\pm0.01}$ & $39.49_{\pm0.29}$ & $21.91_{\pm1.29}$ & $96.57_{\pm0.02}$ & $99.25_{\pm0.08}$          \\ \hline
CB & $64.86_{\pm1.05}$ & $45.83_{\pm2.26}$ & $96.94_{\pm0.07}$ & $99.61_{\pm0.02}$ & $48.16_{\pm1.54}$ & $32.24_{\pm0.94}$ & $96.97_{\pm0.09}$ & $99.32_{\pm0.10}$          \\ \hline
FT & $70.89_{\pm1.93}$ & - & $89.28_{\pm0.22}$ & - & $\textbf{84.31}_{\pm1.32}$ & - & $90.35_{\pm0.15}$ & -          \\ \hline
QDL & $60.87_{\pm1.07}$ & $\textbf{95.21}_{\pm3.09}$ & $89.69_{\pm0.13}$ & $96.72_{\pm0.03}$ & $55.35_{\pm0.75}$ & $\textbf{98.54}_{\pm1.18}$ & $89.72_{\pm0.13}$ & $96.64_{\pm0.13}$          \\ \hline
OSFG & $67.37_{\pm0.64}$ & $50.59_{\pm1.22}$ & $97.12_{\pm0.04}$ & $99.60_{\pm0.03}$ & $56.59_{\pm1.30}$ & $37.29_{\pm1.42}$ & $97.36_{\pm0.06}$ & $99.47_{\pm0.01}$          \\ \hline
GAI$-$ & $70.13_{\pm0.28}$ & $55.70_{\pm0.91}$ & $97.31_{\pm0.02}$ & $99.64_{\pm0.04}$ & $47.47_{\pm2.05}$ & $30.78_{\pm2.75}$ & $96.93_{\pm0.08}$ & $99.29_{\pm0.03}$          \\ \hline
GAI & $\textbf{73.39}_{\pm0.95}$ & $56.68_{\pm1.51}$ & $\textbf{97.56}_{\pm0.08}$ & $\textbf{99.68}_{\pm0.03}$ & $60.78_{\pm0.37}$ & $42.06_{\pm0.89}$ & $\textbf{97.52}_{\pm0.02}$ & $\textbf{99.50}_{\pm0.01}$          \\ \hline
\end{tabular}
}
\label{tbl:benchmark_sub2_10shot}
\end{subtable}
\label{tbl:benchmark_10shot}
\cutAfterTableCaption
\end{table*}

\begin{table*}[!t]
\centering
\cutBeforeTableCaption
\caption{\textbf{Comparison Results on Established $100$-shot Benchmark.} Compared to other state-of-the-art methods, our GAI achieves the best performances, and is robust to choices of majority and minority forgery approaches. }
\cutAfterTableCaption
\begin{subtable}[h]{0.9\textwidth}
\centering
\cutBeforeSubTableCaption
\caption{$100$-shot Benchmarking results on Group1\_FSG and Group1\_DFL.}
\cutAfterSubTableCaption
\scalebox{0.8}{
\begin{tabular}{c|c|c|c|c|c|c|c|c}
\hline
Dataset & \multicolumn{4}{c}{Group1\_FSG} \vline & \multicolumn{4}{c}{Group1\_DFL} \\ \hline
Metric & $\text{ACC}_{\text{minor}}$      & TPR$^{0.1\%}_{\text{minor}}$       & $\text{ACC}_{\text{all}}$      & AUC      & $\text{ACC}_{\text{minor}}$      & TPR$^{0.1\%}_{\text{minor}}$       & $\text{ACC}_{\text{all}}$      & AUC     \\ \hline \hline
Unseen & $26.61_{\pm3.32}$ & $30.01_{\pm2.48}$ & $94.79_{\pm0.24}$ & $98.33_{\pm0.45}$ & $46.10_{\pm10.85}$ & $48.80_{\pm10.14}$ & $95.91_{\pm0.79}$ & $98.89_{\pm0.38}$          \\ \hline
IB & $43.84_{\pm0.61}$ & $53.89_{\pm0.38}$ & $95.97_{\pm0.03}$ & $99.80_{\pm0.01}$ & $41.04_{\pm0.93}$ & $77.89_{\pm1.61}$ & $95.56_{\pm0.07}$ & $99.94_{\pm0.00}$          \\ \hline
CB & $80.82_{\pm2.01}$ & $81.26_{\pm2.21}$ & $98.44_{\pm0.13}$ & $99.70_{\pm0.04}$ & $95.60_{\pm0.90}$ & $95.75_{\pm0.90}$ & $99.42_{\pm0.07}$ & $99.96_{\pm0.01}$          \\ \hline
FT & $62.21_{\pm1.59}$ & - & $87.68_{\pm1.41}$ & - & $89.62_{\pm0.44}$ & - & $91.07_{\pm0.92}$ & -          \\ \hline
QDL & $64.02_{\pm1.87}$ & $\textbf{96.69}_{\pm3.57}$ & $90.70_{\pm0.32}$ & $97.03_{\pm0.24}$ & $40.69_{\pm1.72}$ & $97.11_{\pm1.54}$ & $89.04_{\pm0.06}$ & $96.13_{\pm0.30}$          \\ \hline
OSFG & $84.14_{\pm0.64}$ & $84.79_{\pm1.11}$ & $98.69_{\pm0.04}$ & $99.80_{\pm0.01}$ & $95.22_{\pm0.61}$ & $96.11_{\pm0.42}$ & $99.43_{\pm0.05}$ & $99.96_{\pm0.00}$          \\ \hline
GAI$-$ & $83.17_{\pm1.40}$ & $83.77_{\pm0.52}$ & $98.61_{\pm0.09}$ & $99.80_{\pm0.02}$ & $94.94_{\pm0.81}$ & $95.48_{\pm0.71}$ & $99.40_{\pm0.06}$ & $99.96_{\pm0.00}$          \\ \hline
GAI & $\textbf{85.24}_{\pm0.53}$ & $86.02_{\pm1.03}$ & $\textbf{98.75}_{\pm0.03}$ & $\textbf{99.82}_{\pm0.01}$ & $\textbf{96.85}_{\pm0.30}$ & $\textbf{97.18}_{\pm0.35}$ & $\textbf{99.53}_{\pm0.04}$ & $\textbf{99.97}_{\pm0.00}$          \\ \hline
\end{tabular}
}
\label{tbl:benchmark_sub1_100shot}
\end{subtable}
\begin{subtable}[h]{0.9\textwidth}
\centering
\cutBeforeSubTableCaption
\caption{$100$-shot Benchmarking results on FR+FE\_FSG and FR+FS\_SG.}
\cutAfterSubTableCaption
\scalebox{0.8}{
\begin{tabular}{c|c|c|c|c|c|c|c|c}
\hline
Dataset & \multicolumn{4}{c}{FR+FE\_FSG} \vline & \multicolumn{4}{c}{FR+FS\_SG} \\ \hline
Metric & $\text{ACC}_{\text{minor}}$      & TPR$^{0.1\%}_{\text{minor}}$       & $\text{ACC}_{\text{all}}$      & AUC      & $\text{ACC}_{\text{minor}}$      & TPR$^{0.1\%}_{\text{minor}}$       & $\text{ACC}_{\text{all}}$      & AUC     \\ \hline \hline
Unseen & $59.98_{\pm2.25}$ & $43.58_{\pm5.83}$ & $96.62_{\pm0.10}$ & $99.53_{\pm0.04}$ & $39.90_{\pm1.67}$ & $19.01_{\pm0.78}$ & $96.54_{\pm0.11}$ & $99.15_{\pm0.02}$         \\ \hline
IB & $62.04_{\pm2.23}$ & $36.67_{\pm2.87}$ & $96.79_{\pm0.16}$ & $99.75_{\pm0.01}$ & $54.04_{\pm1.70}$ & $20.37_{\pm2.83}$ & $97.22_{\pm0.10}$ & $99.71_{\pm0.02}$          \\ \hline
CB & $88.76_{\pm0.65}$ & $80.62_{\pm0.83}$ & $98.59_{\pm0.06}$ & $99.82_{\pm0.01}$ & $81.25_{\pm1.25}$ & $70.69_{\pm3.49}$ & $98.47_{\pm0.07}$ & $99.74_{\pm0.01}$          \\ \hline
FT & $85.68_{\pm1.58}$ & - & $90.22_{\pm0.22}$ & - & $\textbf{93.08}_{\pm0.30}$ & - & $90.07_{\pm0.31}$ & -          \\ \hline
QDL & $62.01_{\pm0.65}$ & $\textbf{91.16}_{\pm3.55}$ & $89.66_{\pm0.13}$ & $96.76_{\pm0.16}$ & $56.38_{\pm0.67}$ & $\textbf{99.29}_{\pm0.26}$ & $90.05_{\pm0.03}$ & $97.02_{\pm0.04}$          \\ \hline
OSFG & $89.48_{\pm0.76}$ & $82.57_{\pm1.54}$ & $98.62_{\pm0.06}$ & $99.84_{\pm0.01}$ & $80.72_{\pm0.69}$ & $67.05_{\pm1.38}$ & $98.43_{\pm0.03}$ & $99.75_{\pm0.01}$          \\ \hline
GAI$-$ & $90.17_{\pm0.13}$ & $82.18_{\pm0.86}$ & $98.68_{\pm0.02}$ & $\textbf{99.85}_{\pm0.01}$ & $79.41_{\pm2.23}$ & $66.57_{\pm3.91}$ & $98.40_{\pm0.09}$ & $99.76_{\pm0.01}$          \\ \hline
GAI & $\textbf{90.75}_{\pm0.27}$ & $84.51_{\pm0.64}$ & $\textbf{98.72}_{\pm0.03}$ & $99.84_{\pm0.01}$ & $83.81_{\pm1.05}$ & $73.33_{\pm2.29}$ & $\textbf{98.58}_{\pm0.06}$ & $\textbf{99.79}_{\pm0.02}$          \\ \hline
\end{tabular}
}
\label{tbl:benchmark_sub2_100shot}
\end{subtable}
\label{tbl:benchmark_100shot}
\cutAfterTableCaption
\end{table*}

In this section, we try $10$-shot and $100$-shot settings for the minority class.
We evaluate performances on these two modified benchmarks respectively.

As shown in Tab.~\ref{tbl:benchmark_10shot} where only $10$ minority samples are used, $\text{ACC}_{\text{minor}}$ is mostly around $50\%$ which is not far from the result of blind guesses.
Meanwhile, the standard deviation is also larger than that of 50-shot setting in the main paper, causing the results hard to be statistically significant.
Therefore, $10$ minority samples are not enough for estimating the distribution of the minority class. 

For $100$-shot setting in Tab.~\ref{tbl:benchmark_100shot}, it is not surprising to see significant improvements on all metrics compared to $50$-shot setting in the main paper.
However, no significant reduction is observed in the standard deviation of metrics, which suggests statistical stability is not improved.
In this case, we prefer to choose fewer minority samples, \ie $50$-shot, in our standard benchmark.

Our GAI also achieves the best performances in these two settings.
Similar to 50-shot setting, FT gains the highest $\text{ACC}_{\text{minor}}$ on FR+FS\_SG by simply making more predictions of ``fake", causing a large drop on $\text{ACC}_{\text{all}}$.
Thus, the overall performance of FT is significantly lower than that of GAI.

It is also worth mentioning that GAI brings a larger improvement in the setting with fewer minority samples. 
On $100$-shot setting, performances of SOTAs begin to be close.
It indicates that transferred information will play a more important role when there is less information in the minority domain.

\begin{figure}[!tbh]
\begin{center}
\includegraphics[width=0.9\linewidth]{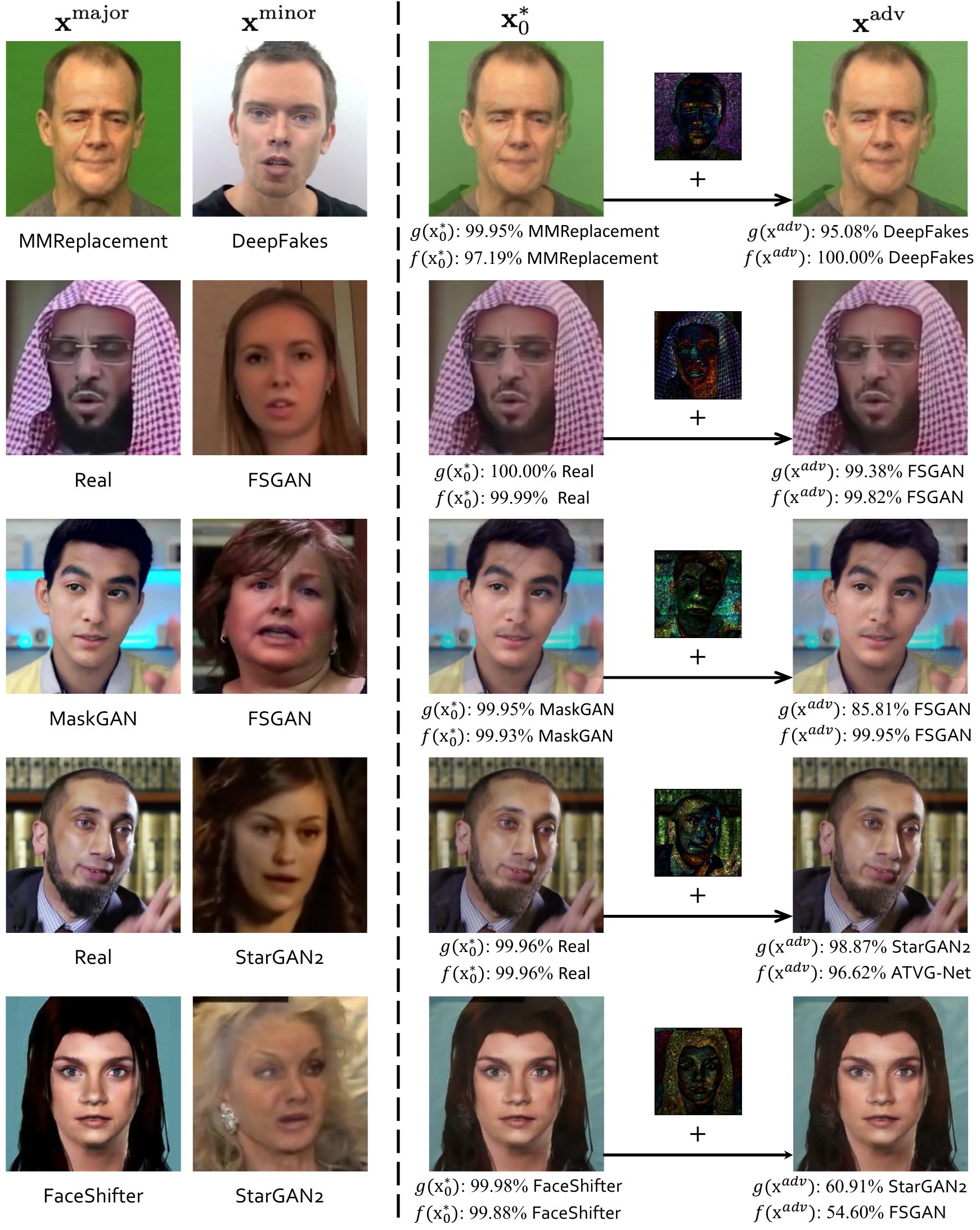}
\end{center}
\cutBeforeFigureCaption
\caption{\textbf{Visualization of Optimization Process.} 
$\mathbf{x}_0^*$ denotes initially generated $\mathbf{x}^*$ with the interpolation coefficient tensor $\pmb{\alpha}$ filled with a constant initialization value $\alpha_0$,
and $\mathbf{x}^{\text{adv}}$ refers to the final generated $\mathbf{x}^*$ after iterative updates. 
}
\cutAfterFigureCaption
\label{supp_fig:diff}
\end{figure}

\subsection{Computational Complexity}
In the inference stage, our method GAI only uses the student model $f$ and thus has no difference from the general neural network.
During the training stage, GAI only performs adversarial interpolation when images from the minority class are sampled, whose probability is around $\frac{1}{m}$. The adversarial interpolation involves forward and backward processing through additional $T$ ($T=10$ in this paper) iterations. However, the computational complexity of remaining $\frac{m-1}{m}$ sampling cases will not be influenced.
As a result, the computational complexity of a single epoch of GAI training is approximately equivalent to three epochs of standard training.

\subsection{Optimization Process.} In Fig.~\ref{supp_fig:diff}, we provide some generated samples and their prediction results to illustrate the changes before and after the iterative optimization ($\mathbf{x}^{\text{adv}}$ versus $\mathbf{x}_0^*$). We can see that if we only use the constant value $\alpha_0$, the interpolated sample will be faithfully recognized as the original forgery class by both teacher net $g$ and student net $f$. However, after adversarial optimization, both teacher net $g$ and student net $f$ will recognize the final result as the new class even if it still belongs to the old class in human eyes. Due to different objective functions for $g$ and $f$, their prediction results may be different. The residual map shows that GAI mainly modifies the attributes and background, which also exhibits that the recognition network is more sensitive to those areas.

\begin{figure}[!tbh]
\begin{center}
\includegraphics[width=0.9\linewidth]{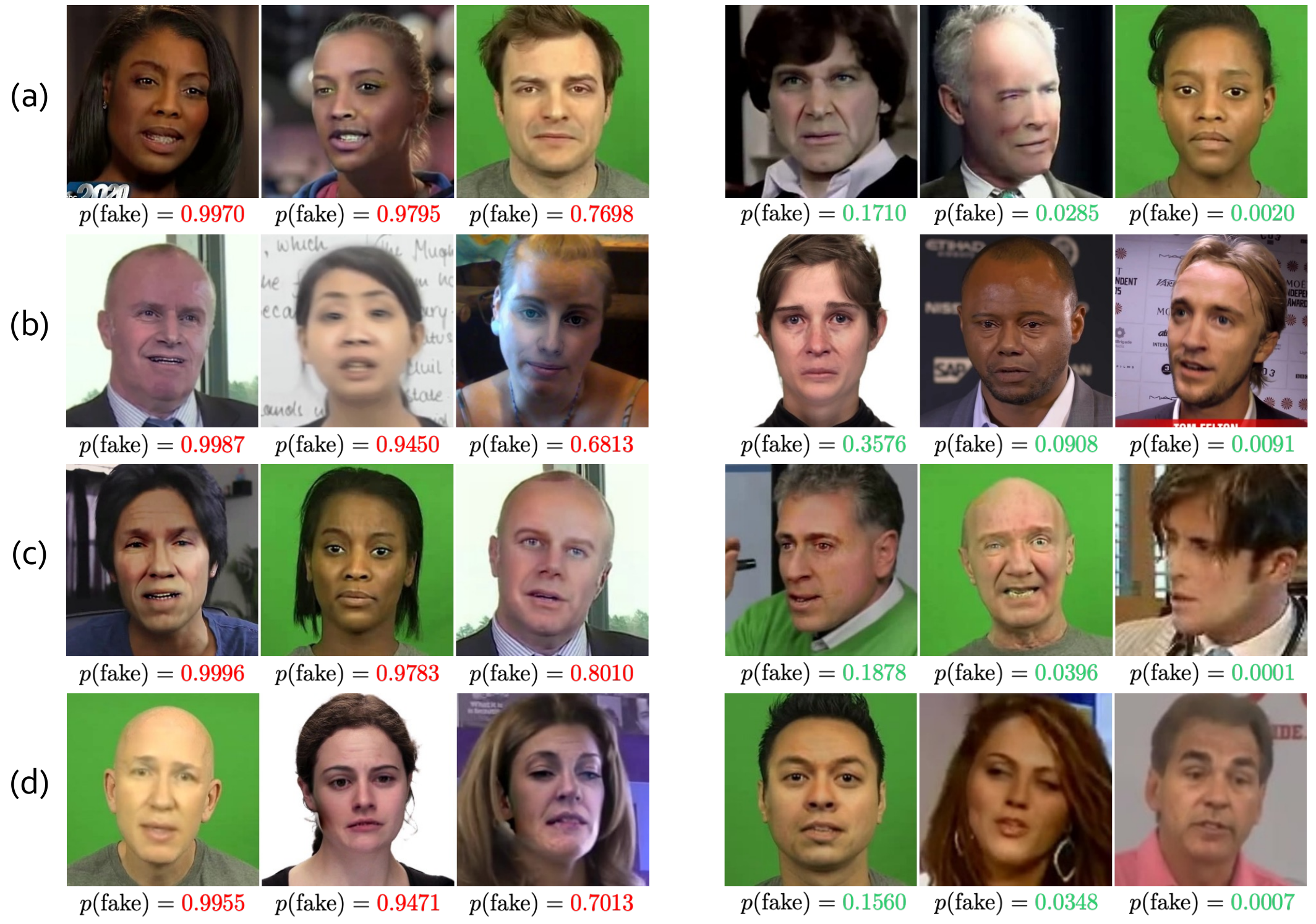}
\end{center}
\cutBeforeFigureCaption
\caption{\textbf{Visualization of Success/Failure Cases.}
(a) Group1\_FSG. (b) Group1\_DFL. (c) FR+FE\_FSG. (d) FR+FS\_SG. The presented samples are all from the test set of the minority class, and the numbers are predicted by our proposed GAI.
}
\cutAfterFigureCaption
\label{supp_fig:failure}
\end{figure}

\subsection{Case Study.} In Fig.~\ref{supp_fig:failure}, we show some success and failure cases. 
For our humans, it is easy to figure out that a man with two different size eyes is forged. But GAI fails to detect a generated man who opens one eye but closes the other with crazy expressions. That may be due to GAI never seeing similar cases during the training stage.
However, most forgery samples look like real pictures at first glance, which requires the detection model to extract imperceptible forged features.
In the current stage, we still can not conclude the general reason for failure cases.

\cutBeforeSection
\section{Conclusion}
\cutAfterSection

In this work, we consider the few-shot forgery detection task and put forward a large-scale benchmark for evaluation.
To the best of our knowledge, this is the first benchmark for few-shot forgery detection with comprehensive forgery approach combinations.
Ideally, a robust forgery detection method should be capable of detecting forged samples without prior knowledge of their forgery approaches (zero-shot setting). However, current technologies have only been able to detect several typical forgery types, and there is a need to continue using the few-shot setting in this benchmark. 
This means that defenders can collect newly forged samples that successfully exploit their systems and update their models to detect new forgery types, which is also a common scenario in the real world.
Forgery approaches in our benchmark include face transfer, face swap, face reenactment, and face editing, which cover most types of forgery types. As a result, our benchmark is useful to evaluate whether a robust forgery detection method is effective for various forgery types or only for specific ones.

Meanwhile, we propose a concise yet effective framework, Guided Adversarial Interpolation (GAI), that adversarially interpolates forgery artifacts of minority samples with majority samples under the guidance of a teacher network.
Extensive experiments demonstrate that our method achieves state-of-the-art performances on the established benchmark.
This notable improvement in performance can be attributed to the integration of valuable information from both the minority and majority domains.
Our work provides a useful stepping stone for evaluating few-shot forgery detection methods, and will hopefully encourage further development in this field.

Finally, there are still some limitations in our work. While creating our few-shot detection benchmark, we analyzed various forgery approaches and found that certain approaches had correlations with each other. For example, a model trained on samples generated by forgery approach A may perform well in detecting forgery approach B, but poorly for forgery approach C. This suggests that certain forgery approaches may share similar characteristics. To further investigate this, we grouped the forgery methods together based on these similarities. However, we still need to understand how these correlations develop within each group. 
In future work, we plan to dig out the shared forgery traits among approaches with strong correlations and design more effective forgery detection methods for each type of forgery approach based on these shared traits.


\cutBeforeSection
\section{Acknowledgements}
\cutAfterSection

This research/project is supported by the National Research Foundation, Singapore
 under its AI Singapore Programme (AISG Award No: AISG2-PhD-2022-01-035T), NTU NAP, MOE AcRF Tier 1 (2021-T1-001-088), and
under the RIE2020 Industry Alignment Fund – Industry Collaboration Projects (IAF-ICP) Funding Initiative, as well as cash and
in-kind contribution from the industry partner(s).

\cutBeforeSection



\bibliographystyle{elsarticle-num}
\setlength{\bibsep}{0.5ex}
\bibliography{egbib}






\newpage

\begin{wrapfigure}{l}{25mm} 
    \includegraphics[width=1in,height=1.25in,clip,keepaspectratio]{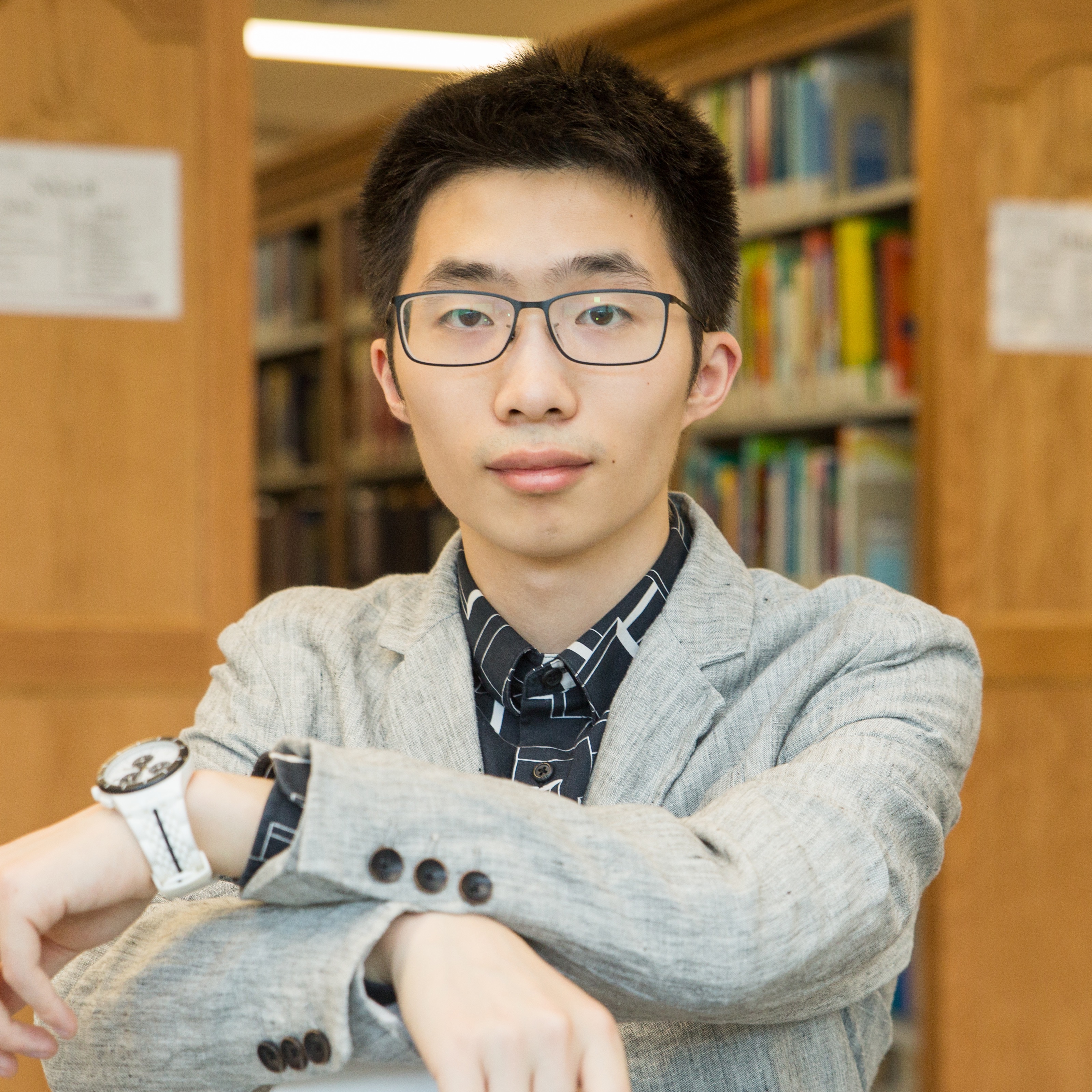}
\end{wrapfigure}\par
\textbf{Haonan Qiu} is a PhD student at NTU SCSE department, supervised by Prof. Ziwei Liu. He received the bachelor degree from The Chinese University of Hong Kong, Shenzhen, majoring in computer science at School of Science and Engineering. His research interests mainly focus on Deep Generative Models, Adversarial Machine Learning.\par

\begin{wrapfigure}{l}{25mm} 
    \includegraphics[width=1in,height=1.25in,clip,keepaspectratio]{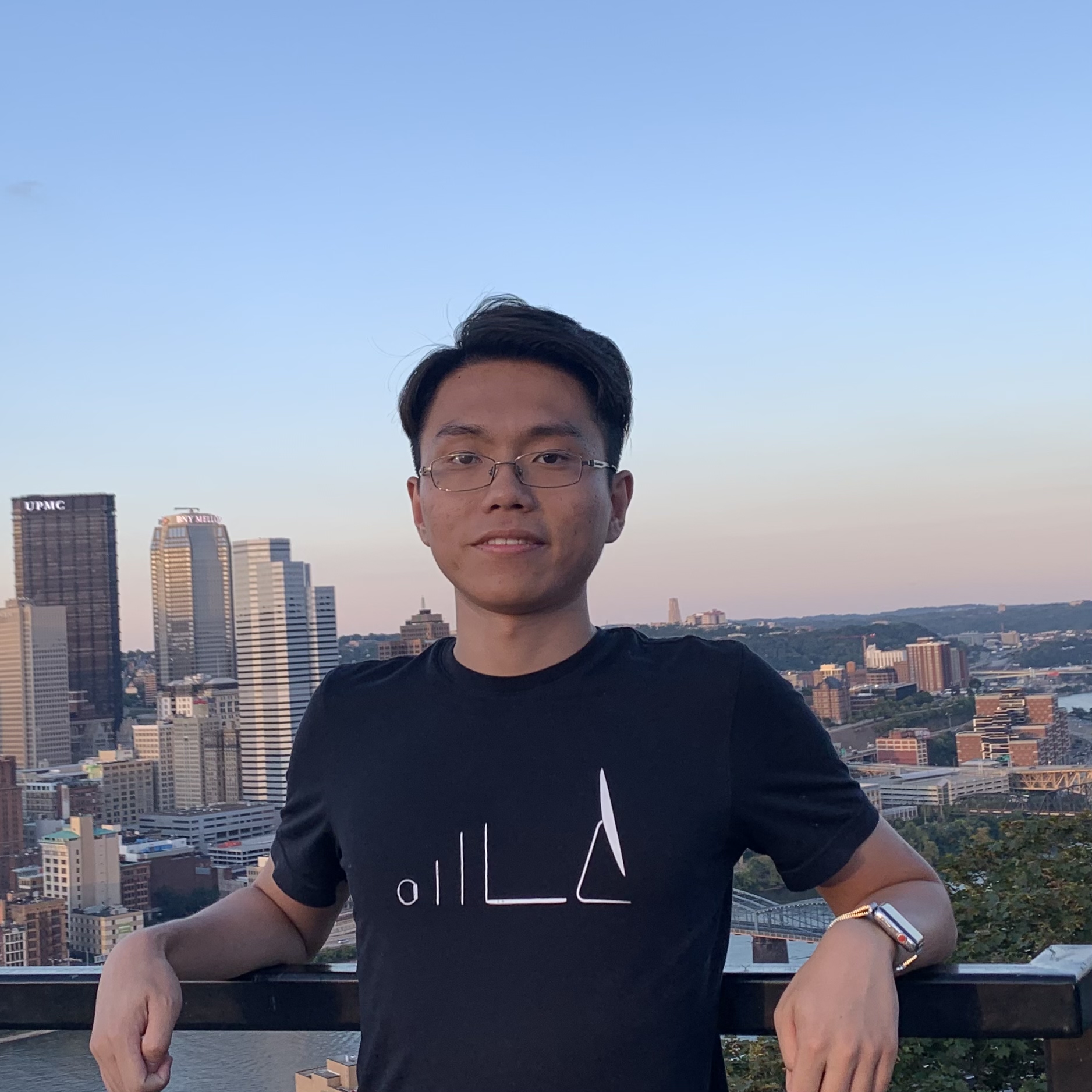}
\end{wrapfigure}\par
\textbf{Siyu Chen} is currently a Master’s student at Carnegie Mellon University majoring in machine learning. He did an internship at SenseTime Research starting from 2019, under the supervision of Dr. Jing Shao. His research interest mainly lies in deep learning and computer vision.\par

\begin{wrapfigure}{l}{25mm} 
    \includegraphics[width=1in,height=1.25in,clip,keepaspectratio]{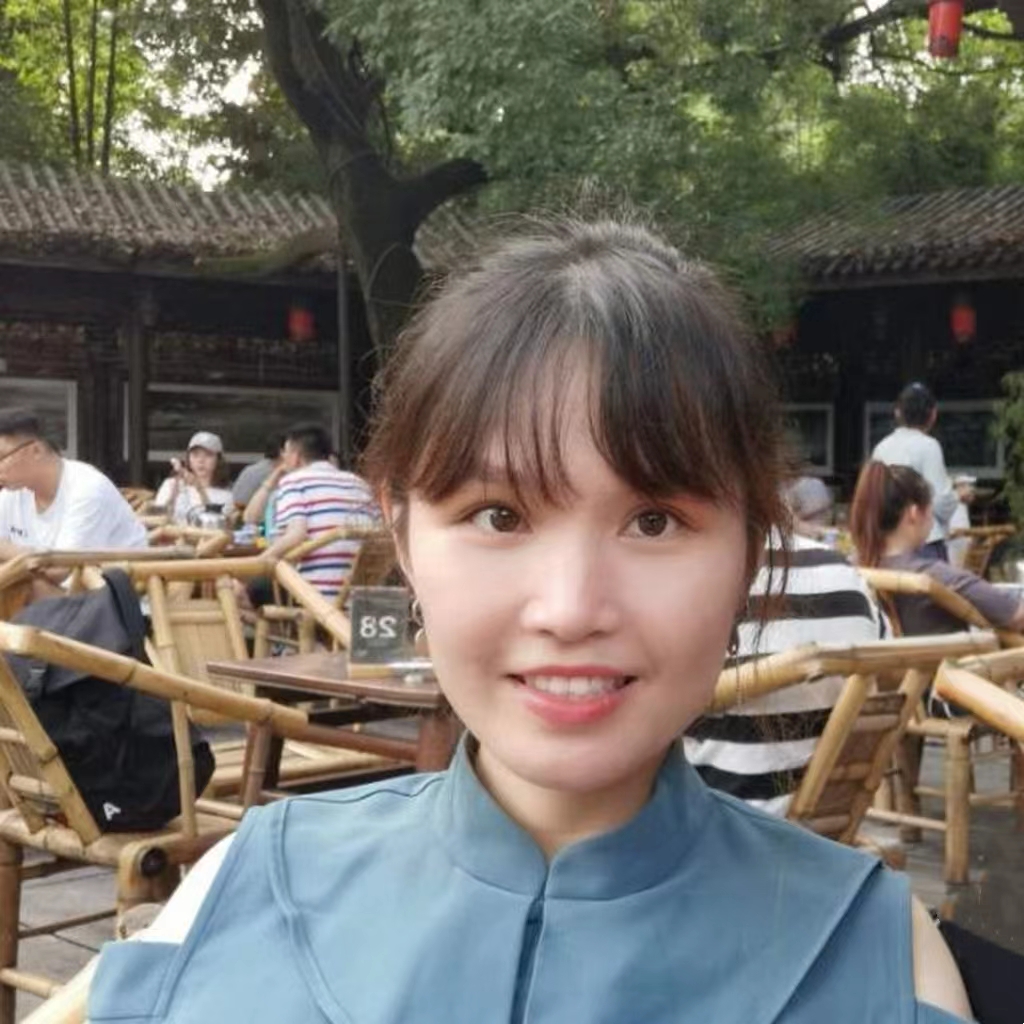}
\end{wrapfigure}\par
\textbf{Bei Gan} was a computer vision researcher at sensetime. Bei Gan received a master's degree from Beijing University of Posts and Telecommunications. At sensetime, her research interests mainly focus on anti-spoofing detection, forgery detection and deep generative model.\par

\begin{wrapfigure}{l}{25mm} 
    \includegraphics[width=1in,height=1.25in,clip,keepaspectratio]{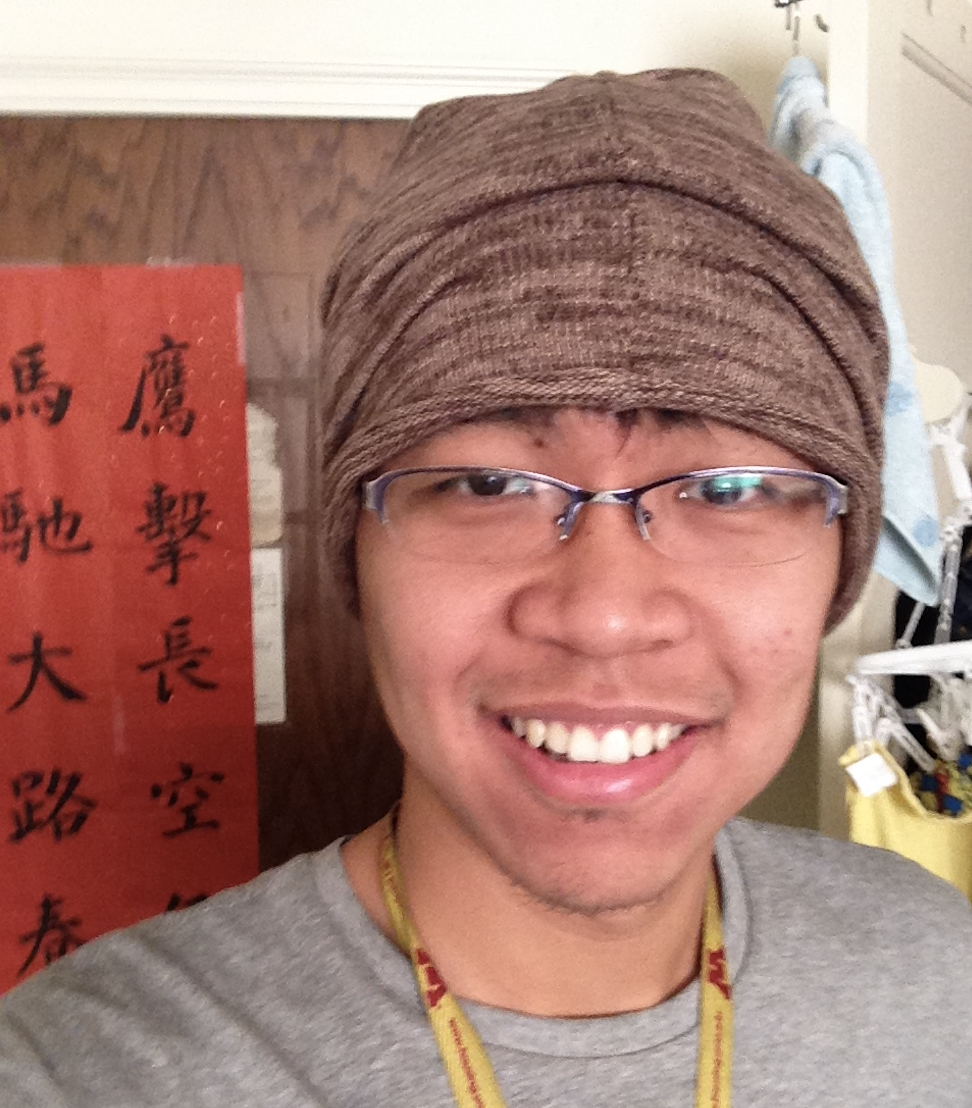}
\end{wrapfigure}\par
\textbf{Kun Wang} is currently a senior researcher at SenseTime Group Limited. Kun Wang received the Mphil degree from the Department of Electronic Engineering, Chinese University of Hong Kong. His research interests include computer vision and deep learning, especially learning better feature representations. \par

\newpage

\begin{wrapfigure}{l}{25mm} 
    \includegraphics[width=1in,height=1.25in,clip,keepaspectratio]{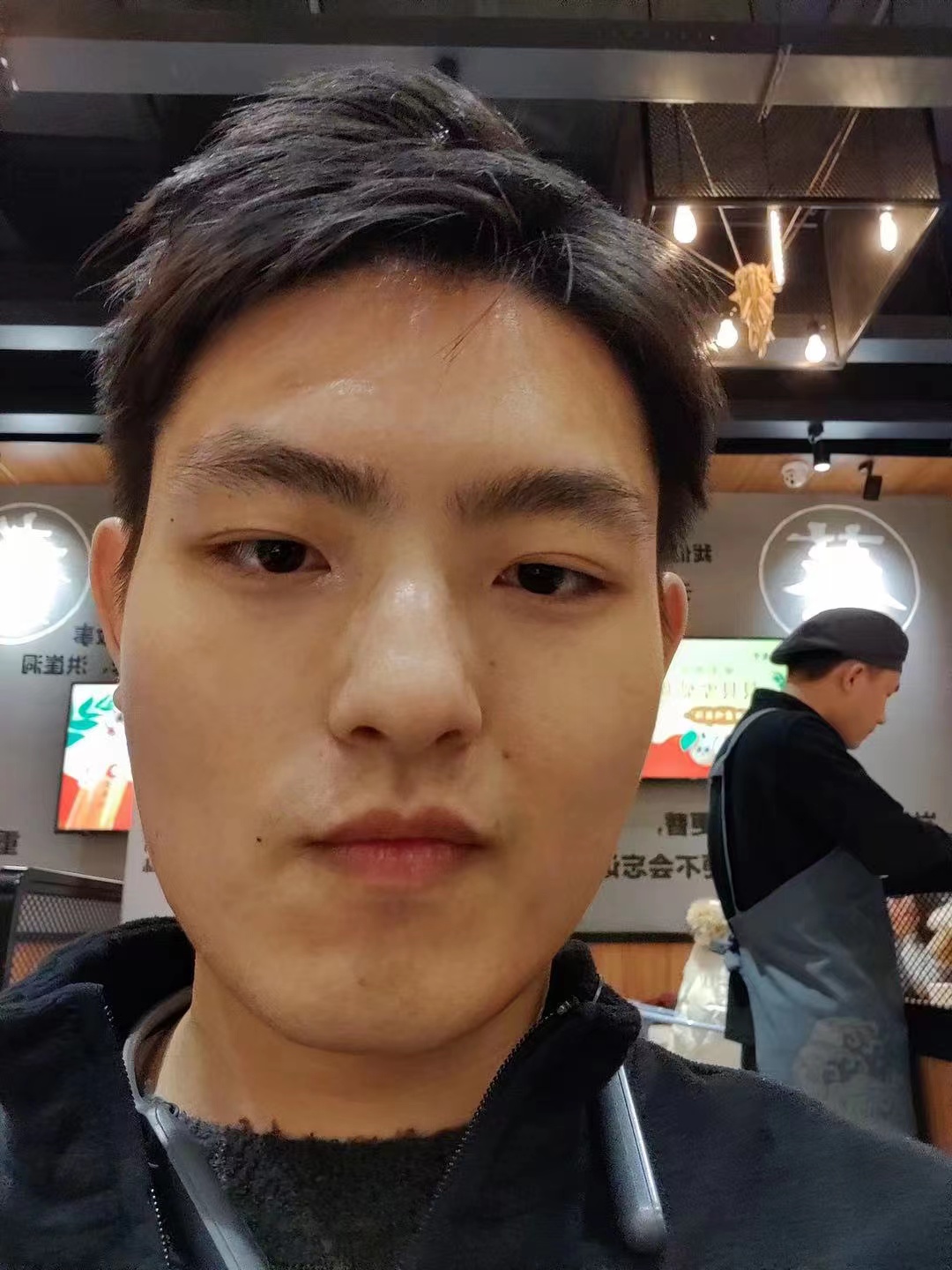}
\end{wrapfigure}\par
\textbf{Huafeng Shi} is currently a computer vision researcher at Sensetime Research. His current research interests include deepfake, face anti-spoofing and data poisoning. He received the bachelor‘s degree in software engineering from Hebei Normal University (2015) and the master's degree in computer technology from Beijing University of Posts and Telecommunications (2020).\par

\begin{wrapfigure}{l}{25mm} 
    \includegraphics[width=1in,height=1.25in,clip,keepaspectratio]{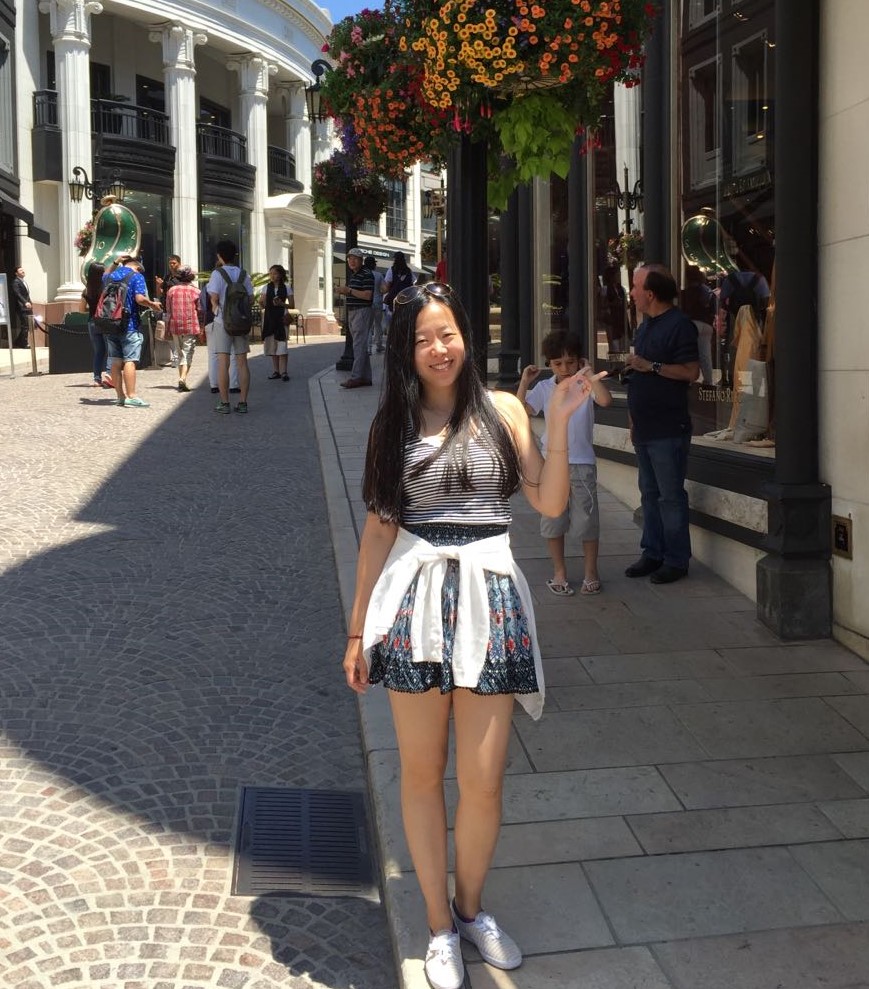}
\end{wrapfigure}\par
\textbf{Jing Shao} is currently a Vice Director in SenseTime Group Limited. She received her PhD (2016) in Electronic Engineering from The Chinese University of Hong Kong (CUHK), supervised by Prof. Xiaogang Wang, and work closely with Prof. Chen Change Loy and the Multimedia Lab (MMLab) led by Prof. Xiaoou Tang.\par

\begin{wrapfigure}{l}{25mm} 
    \includegraphics[width=1in,height=1.25in,clip,keepaspectratio]{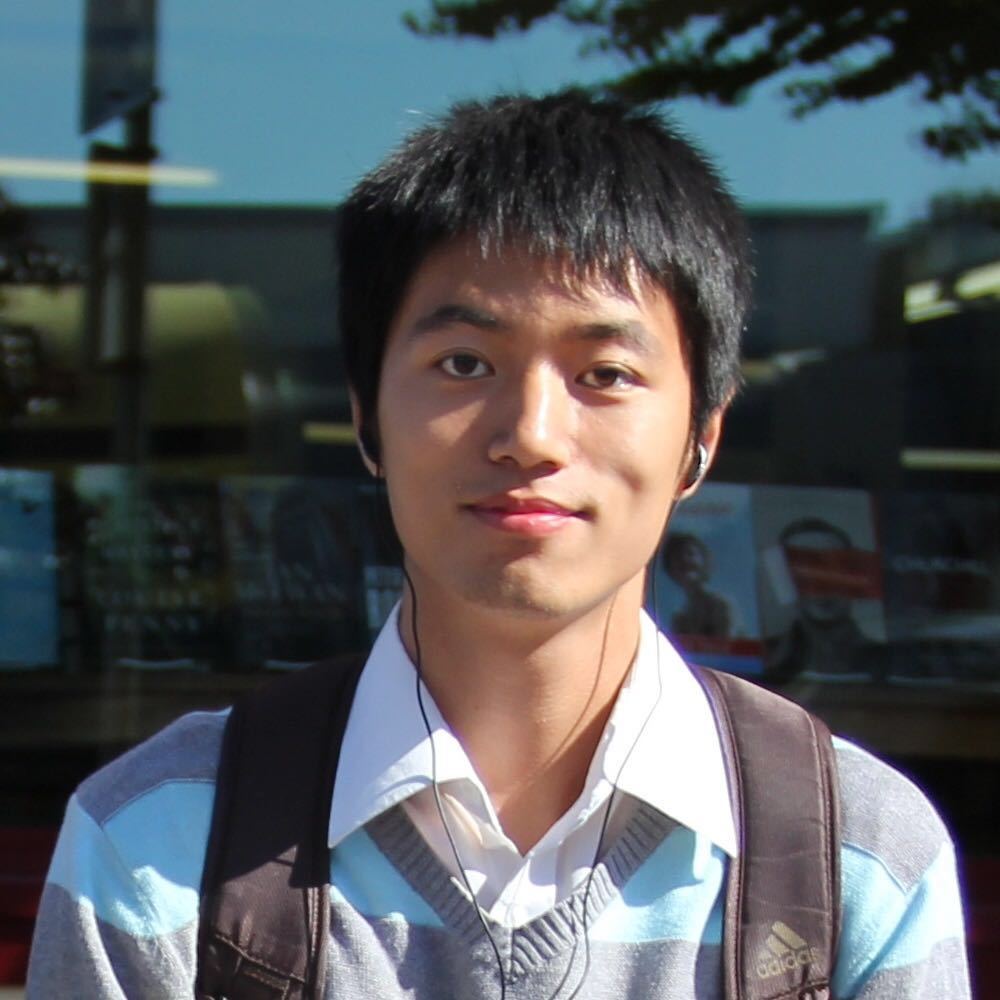}
\end{wrapfigure}\par
\textbf{Ziwei Liu} is currently a Nanyang Assistant Professor at Nanyang Technological University. Previously, Ziwei was a postdoctoral researcher at University of California, Berkeley, working with Prof. Stella Yu. Ziwei received his PhD from the Chinese University of Hong Kong in 2017, supervised by Prof. Xiaoou Tang and Prof. Xiaogang Wang.\par

\end{document}